\documentclass[10pt,journal,compsoc]{IEEEtran}

\usepackage{times}
\usepackage{epsfig}
\usepackage{graphicx}
\usepackage{amsmath}
\usepackage{amsthm}
\usepackage{amssymb}
\usepackage{dsfont}
\usepackage{algorithmic}
\usepackage{stfloats}
\usepackage{color}
\usepackage{soul}
\usepackage{algorithm,algorithmic}
\usepackage{subcaption}

\usepackage{siunitx}
\usepackage{booktabs}

\usepackage[pagebackref=false,breaklinks=true,letterpaper=true,colorlinks,bookmarks=false]{hyperref}
\newif\ifdraft
\draftfalse
\drafttrue
\usepackage{color}

\definecolor{orange}{rgb}{1,0.5,0}
\definecolor{violet}{RGB}{70,0,170}
\definecolor{brown}{rgb}{0.48, 0.25, 0.0}
\definecolor{darkgreen}{rgb}{0, 0.7, 0}
\definecolor{aqua}{rgb}{0.19, 0.55, 0.91}
\ifdraft
 \newcommand{\PF}[1]{{\color{red}{\bf PF: #1}}}
 
 \newcommand{\KY}[1]{{\color{blue}{\bf KY: #1}}}
 \newcommand{\ky}[1]{{\color{blue} #1}}
 \newcommand{\MS}[1]{{\color{darkgreen}{\bf MS: #1}}}
 
 \newcommand{\ZD}[1]{{\color{violet}{\bf ZD: #1}}}
 \newcommand{\zd}[1]{{\color{violet}{#1}}}
 \newcommand{\YH}[1]{{\color{orange}{\bf YH: #1}}}
 
 \newcommand{\ok}[1]{{\color{black}{#1}}}
\else
 \newcommand{\PF}[1]{}
 
 \newcommand{\KY}[1]{}
 \newcommand{\ky}[1]{ #1 }
 \newcommand{\MS}[1]{}
 
 \newcommand{\ZD}[1]{}
 \newcommand{\zd}[1]{{#1}}
 \newcommand{\YH}[1]{}
 
\fi

\newcommand{\comment}[1]{}

\newcommand{\bA}{\mathbf{A}}
\newcommand{\bU}{\mathbf{U}}
\newcommand{\bS}{\mathbf{S}}
\newcommand{\bM}{\mathbf{M}}
\newcommand{\bK}{\mathbf{K}}
\newcommand{\bR}{\mathbf{R}}

\newcommand{\bbA}{\bar{\mathbf{A}}}
\newcommand{\bW}{\mathbf{W}}
\newcommand{\bX}{\mathbf{X}}
\newcommand{\bxi}{\boldsymbol{\xi}}
\newcommand{\btheta}{\boldsymbol{\theta}}

\newcommand{\bXe}{\tilde{\mathbf{X}}}
\newcommand{\bbX}{\bar{\mathbf{X}}}
\newcommand{\bI}{\mathbf{I}}

\newcommand{\bw}{\mathbf{w}}
\newcommand{\bq}{\mathbf{q}}
\newcommand{\be}{\mathbf{e}}
\newcommand{\bte}{\tilde{\mathbf{e}}}
\newcommand{\bx}{\mathbf{x}}

\newcommand{\bt}{\mathbf{t}}

\def\bSigma{\boldsymbol\Sigma}

\newcommand{\minimize}{\operatornamewithlimits{minimize}}

\newcommand{\Fig}[1]{Fig.~\ref{fig:#1}}
\newcommand{\Eq}[1]{Eq.~\eqref{eq:#1}}

\begin{document}

\author{Zheng~Dang, Kwang~Moo~Yi, Yinlin~Hu, Fei~Wang, Pascal~Fua,~\IEEEmembership{Fellow,~IEEE}, and~Mathieu~Salzmann
\\
	\IEEEcompsocitemizethanks{
		\IEEEcompsocthanksitem Z. Dang is with the National Engineering Laboratory for Visual Information Processing and Application, Xi’an Jiaotong University, Xi'an, China.
		\protect\\
		E-mail: dangzheng713@stu.xjtu.edu.cn
		\IEEEcompsocthanksitem K. M. Yi is with the Visual Computing Group, University of Victoria, Canada.
		\protect\\
		E-mail: kyi@uvic.ca
		\IEEEcompsocthanksitem F. Wang is with the National Engineering Laboratory for Visual Information Processing and Application, Xi’an Jiaotong University, Xi'an, China.
		\protect\\
		E-mail: wfx@mail.xjtu.edu.cn
		\IEEEcompsocthanksitem Y. Hu, P. Fua and M. Salzmann are with the Computer Vision Laboratory, \'{E}cole Polytechnique F\'{e}d\'{e}rale de Lausanne, Lausanne CH-1015, Switzerland.
		\protect\\
		E-mail: firstname.lastname@epfl.ch\protect\\
	}
}

\title{Eigendecomposition-Free Training of Deep Networks for Linear Least-Square Problems}

\IEEEtitleabstractindextext{

\begin{abstract}

Many classical Computer Vision problems, such as essential matrix computation and pose estimation from 3D to 2D correspondences, can be tackled by solving a linear least-square problem, which can be done by finding the eigenvector corresponding to the smallest, or zero, eigenvalue of a matrix representing a linear system. Incorporating this in deep learning frameworks would allow us to explicitly encode known notions of geometry, instead of having the network implicitly learn them from data. However, performing eigendecomposition within a network requires the ability to differentiate this operation. While theoretically doable, this introduces numerical instability in the optimization process in practice. 
In this paper, we introduce an eigendecomposition-free approach to training a deep network whose loss depends on the eigenvector corresponding to a zero eigenvalue of a matrix predicted by the network.
We demonstrate that our approach is much more robust than explicit differentiation of the eigendecomposition using two general tasks, outlier rejection and denoising, with several practical examples including wide-baseline stereo, the perspective-n-point problem, and ellipse fitting. Empirically, our method has better convergence properties and yields state-of-the-art results.

\end{abstract}

}

\maketitle

\section{Introduction}
\label{sec:intro}

\IEEEPARstart{T}{h}e solution of many Computer Vision problems can be obtained by finding the zero singular- or eigen-value of a matrix, that is, by solving a  
linear least-square problem. This is true of model-fitting tasks ranging from plane and ellipse fitting~\cite{Zhang95c,Kanatani16} to wide-baseline stereo~\cite{Hartley00} and the perspective-n-point (PnP) problem~\cite{Lepetit09, Ferraz14,Zheng13, Garro12,Li12c}. With the advent of Deep Learning, there has been a push to embed such least-squares approaches in deep architectures to enforce constraints. For example, we have recently shown that this could be done when training networks to detect and match keypoints in image pairs while accounting for the global epipolar geometry~\cite{Yi18a}. More generally, this approach makes it possible to explicitly incorporate geometrical knowledge into the network~\cite{Ranftl18}. As a result, it operates in a more constrained space, which allows learning from smaller amounts of training data.
 
\begin{figure}[t]
    \centering
    \begin{tabular}{cc}
     \parbox[c]{0.4\linewidth}{\includegraphics[scale=0.1,width=1.0\linewidth]{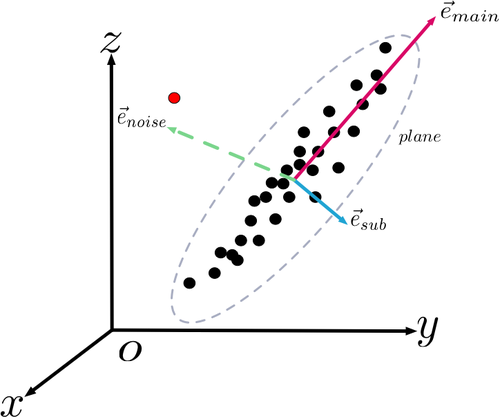}} &
     \parbox[c]{0.48\linewidth}{\includegraphics[scale=0.1,width=1.0\linewidth]{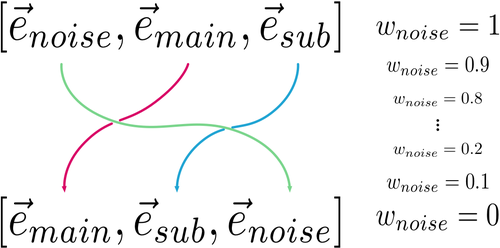}}\\[-2mm]
    (a) & (b)
    \end{tabular}{}
    
 \caption{{\bf Eigenvector switching.}
   {\bf (a)} 3D points lying on a plane in black and distant outlier in red. {\bf (b)} When the weights assigned to all the points are one, the eigenvector corresponding to the smallest eigenvalue is $\be_{sub}$, the vector shown in blue in (a), and on the right in the top portion of (b), where we sort the eigenvectors by decreasing eigenvalue. As the optimization progresses and the weight assigned to the outlier decreases, the eigenvector corresponding to the smallest eigenvalue switches to $\be_{noise}$, the vector shown in green in (a), which introduces a sharp change in the gradient values. }
\label{fig:switching}

\end{figure}

The currently dominant way to implement this approach is to design a network whose output is a matrix and train the network so that the smallest singular- or eigen-vector of that matrix is as close as possible to the ground-truth one~\cite{Yi18a,Ranftl18}.  This requires differentiating the singular value decomposition (SVD) or the eigendecomposition (ED) in a stable manner at training time. This problem has received considerable attention~\cite{Papadopoulo00,Giles08,Ionescu15} and decomposition algorithms are already part of standard Deep Learning frameworks, such as TensorFlow~\cite{Tensorflow} or PyTorch~\cite{PyTorch}. Unfortunately, these algorithms suffer from two main weaknesses. First,  when optimizing with respect to the matrix coefficients or with respect to parameters defining them, the vector corresponding to the smallest singular value or eigenvalue may change abruptly as the relative magnitudes of these values evolve, which is a  non-differentiable behavior. This is illustrated in Fig.~\ref{fig:switching}, discussed in detail in Section~\ref{sec:motivation}, and demonstrated experimentally in Section~\ref{sec:planefitting}. Second, computing the derivatives of the eigenvector with respect to the coefficients of the input matrix requires dividing by the difference between two singular values or eigenvalues, which could be very small. While a solution to the second problem was proposed in~\cite{Papadopoulo00}, the first remains and handicaps all methods that depend on SVD or ED.

In this paper, we therefore introduce a method that can enforce the same global constraints as the SVD- and ED-based methods but without explicitly performing these decompositions, thus eliminating the non-differentiability issue and improving numerical stability. More specifically, given the network weights $\theta$, let $\bA_\theta$ be the matrix that the network outputs and let $\bte$ be the true eigenvector corresponding to the zero eigenvalue. We introduce a loss function that is minimized when $\| \bA_\theta \bte \| = 0$ while disallowing trivial solutions such as $\bA_\theta=\mathbf{0}$. In practice,  image measurements are not perfect and the eigenvalue is never strictly zero. However, this does not significantly affect the computation, which makes our approach robust to noise.

We will demonstrate that our approach delivers better results than using the standard implementation of singular- and eigen-value decomposition provided in TensorFlow for
plane and ellipse fitting, solving the PnP problem, and distinguishing good keypoint correspondences from bad ones. A preliminary version of this paper appeared in~\cite{Dang18a}. In this extended version, we use a toy example to better motivate our approach and show how the gradients behave; we expand the formulation to handle denoising; we perform simultaneous outlier rejection and denoising; we demonstrate the application of our method to ellipse fitting. Note that, while we only discuss our approach in the context of Deep Learning, it is applicable to \textit{any} optimization problem in which one seeks to optimize a loss function based on the smallest eigenvector of a matrix with respect to the parameters that define this matrix.

\section{Motivation}
\label{sec:motivation}

To illustrate the problems associated with differentiating eigenvectors and eigenvalues, consider the outlier rejection example depicted by Fig.~\ref{fig:switching}. The inputs are 3D inlier points lying on a plane and an outlier assumed to be very far from the plane. The inliers are shown in black and the outlier in red. 

Suppose we want to assign a binary weight to each point---1 for inliers, 0 for outliers---such that the eigenvector corresponding to the smallest eigenvalue of the weighted covariance matrix is close to the ground-truth one in the least-square sense. When the weight assigned to the outlier is 0, it would be $\be_{noise}$, which is also the normal to the plane and is shown in green.  However, if at any point during the optimization we assign the weight 1 to the outlier, $\be_{noise}$ will correspond to the largest eigenvalue instead of the smallest one and the eigenvector corresponding to the smallest eigenvalue will be the vector $\be_{sub}$ shown in blue, which is perpendicular to $\be_{noise}$. As a result, if we initially set all weights to 1 and optimize them so that the smallest eigenvector approaches the plane normal, the gradient values will depend on the coordinates of $\be_{sub}$.  If everything goes well, at some point during the optimization, the weight assigned to the outlier will become small enough so that the smallest eigenvector switches from being $\be_{sub}$ to being $\be_{noise}$. 

However, this introduces a large jump in the gradient vector whose values will now depend on the coordinates of $\be_{noise}$ instead of $\be_{sub}$. In this simple case, the resulting instability does not preclude eventual convergence. However, in more complex situations, we found that it does. We already noted this problem in~\cite{Yi18a} when simultaneously establishing keypoint correspondences and estimating the essential matrix. Because of it, we had  to first rely solely on a classification loss  to determine the potential inlier correspondences before incorporating the loss based on the essential matrix to impose geometric constraints, which requires ED. This ensured that the network weights were already good enough to prevent eigenvector switching when starting to minimize the geometry-based loss.
This strategy, however, requires determining ground-truth inlier/outlier labels, which in practice may be difficult to obtain. Furthermore, this scheme may not even be possible in cases where one cannot provide an auxiliary classification task---for example in the traditional task of removing additive noise in data. As we will show later, our method does not suffer from this.

\section{Related Work}

In recent years, the need to integrate geometric methods and mathematical tools into Deep Learning frameworks has led to the reformulation of a number of them in network terms. For example,~\cite{Jaderberg15} considers spatial transformations of image regions with CNNs. The set of such transformations is extended in~\cite{Handa16}. In a different context,~\cite{Murray16} derives a differentiation of the Cholesky decomposition that could be integrated in Deep Learning frameworks.  

Unfortunately, the set of geometric Computer Vision problems that these methods can handle remains relatively limited.
In particular, there is no widely accepted deep-learning way to solve the many geometric problems that reduce to finding the least-square solution to a linear system.
In this work, we consider three such problems: Ellipse fitting, estimating the
3D pose of an object from 3D-to-2D correspondences, and computing the essential matrix from keypoint correspondences in an image pair, all of which we briefly discuss below.\\

\noindent{\bf Ellipse fitting.} 
This is a traditional Computer Vision problem, where one finds the ellipse equation that best fits a set of points.
This could be used, for example, to detect people's eyes in images by fitting ellipses to detected image edges.
One of the simplest and most straightforward way to solve this problem is the Direct Linear Transform (DLT)~\cite{Zhang95c, Kanatani16}.
The DLT corresponds to finding an algebraic least-square solution to the problem by performing ED or SVD and retaining the eigen- or singular-vector that corresponds to the minimum eigen- or singular-value.
Besides the DLT, other methods focus on minimizing the geometric distance instead of the algebraic one~\cite{Zhang95c}, and exploit constraints to enforce the result to truly be an ellipse, as opposed to a general conic, which also encompasses parabolas and hyperbolas~\cite{Fitzgibbon99, Halir98}.
In practice, outliers are handled in an iterative manner, where the solution is refined via robust techniques such as M-estimators~\cite{Zhang95c} or Least Median of Squares (LMedS)~\cite{Rousseeuw87}. For a detailed review of ellipse fitting, we refer the reader to~\cite{Kanatani16}.\\

\noindent{\bf Estimating 3D pose from 3D-to-2D correspondences.} This is known as the Perspective-n-Point (PnP) problem. It has also been investigated for decades and is also amenable to an eigendecomposition-based solution~\cite{Hartley00}, many variations of which have been proposed over the years~\cite{Lepetit09,Kneip11,Zheng13,Ferraz14}.  DSAC~\cite{Brachmann16b} is the only approach we know of that integrates the PnP solver into a Deep Network. As explicitly differentiating through the PnP solver is not optimization friendly, the authors apply the log trick used in the reinforcement learning literature. This amounts to using a numerical approximation of the derivatives from random samples, which is not ideal, given that an analytical alternative exists. Moreover, DSAC only works for grid configurations and known scenes. By contrast, the method we propose in this work has an analytical form, with no need for stochastic sampling.\\

\noindent{\bf Estimating the Essential matrix from correspondences.} The eigenvalue-based solution to this problem has been known for decades~\cite{Longuet-Higgins81,Hartley97,Hartley00} and remains the standard way to compute Essential matrices~\cite{Nister03}. The real focus of research in this area has been to establish reliable keypoint correspondences  and to eliminate outliers. In this context, variations of RANSAC~\cite{Fischler81}, such as MLESAC~\cite{Torr00} and least median of squares (LMeds)~\cite{Rousseeuw87}, and very recently GMS~\cite{Bian17}, have become popular. For a comprehensive study of such methods, we refer the interested reader to~\cite{Raguram13}. With the emergence of Deep Learning, there has been a trend towards moving away from this decades-old knowledge and apply instead a black-box approach where a Deep Network is trained to directly estimate the rotation and translation matrices~\cite{Zamir16,Ummenhofer17} without {\it a priori} geometrical knowledge. Our very recent work~\cite{Yi18a} attempts to reconcile these two opposing trends by embedding the geometric constraints into a Deep Net and has demonstrated superior performance for this task when the correspondences are hard to establish. \ok{A similar approach was also proposed in the contemporary work of~\cite{Ranftl18}.}\\
\begin{figure*}[t]
\centering
\begin{subfigure}{.19\textwidth}
    \centering
    \includegraphics[scale=0.01,width=1.\linewidth]{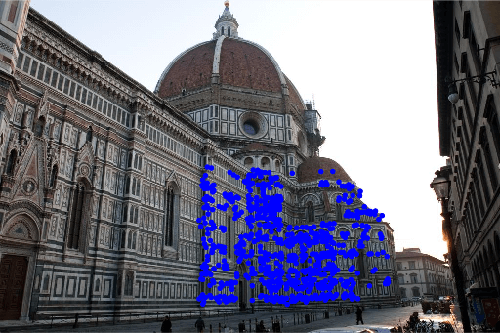}
\end{subfigure}
\begin{subfigure}{.19\textwidth}
    \centering
    \includegraphics[scale=0.01,width=1.\linewidth]{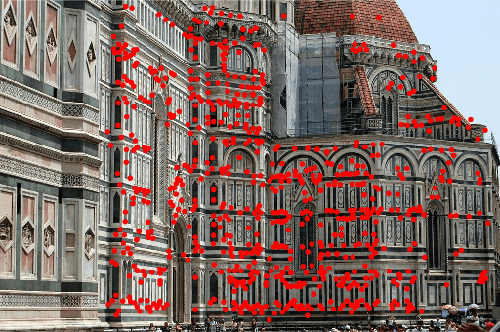}
\end{subfigure}
\begin{subfigure}{.20\textwidth}
    \centering
    \includegraphics[scale=0.01,width=1.\linewidth]{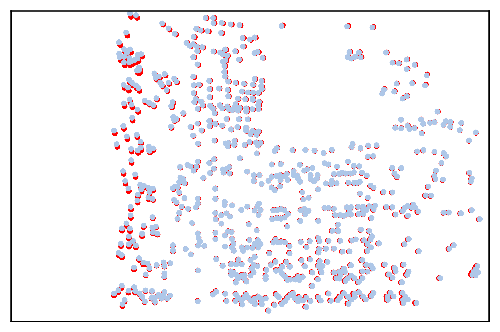}
\end{subfigure}
\begin{subfigure}{.20\textwidth}
    \centering
    \includegraphics[scale=0.01,width=1.\linewidth]{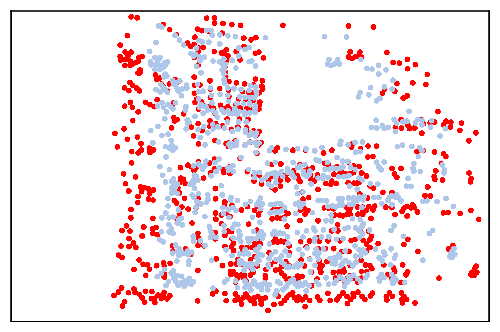}
\end{subfigure}
\begin{subfigure}{.20\textwidth}
    \centering
    \includegraphics[scale=0.01,width=1.\linewidth]{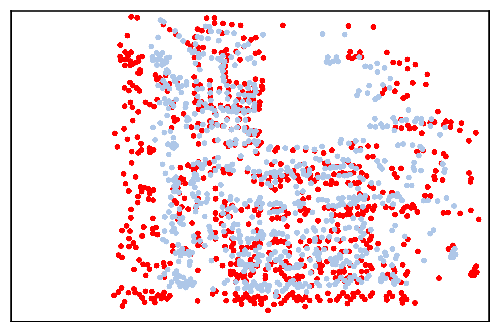}
\end{subfigure}

\begin{subfigure}{.19\textwidth}
	\centering
	\includegraphics[scale=0.01,width=1.\linewidth, height=2.31cm]{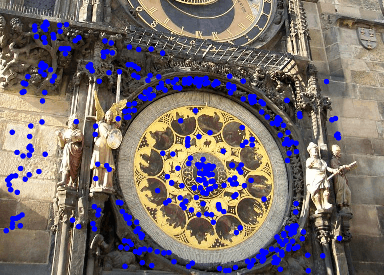}
	\caption*{Base Image}
\end{subfigure}
\begin{subfigure}{.19\textwidth}
	\centering
	\includegraphics[scale=0.01,width=1.\linewidth]{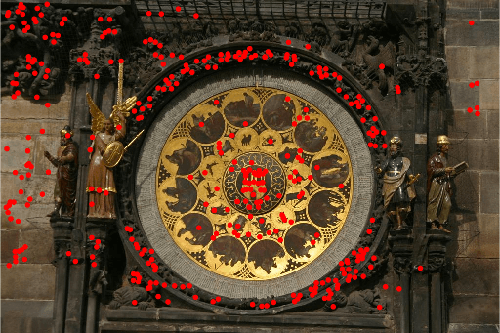}
	\caption*{Matching Image}
\end{subfigure}
\begin{subfigure}{.20\textwidth}
	\centering
	\includegraphics[scale=0.01,width=1.\linewidth]{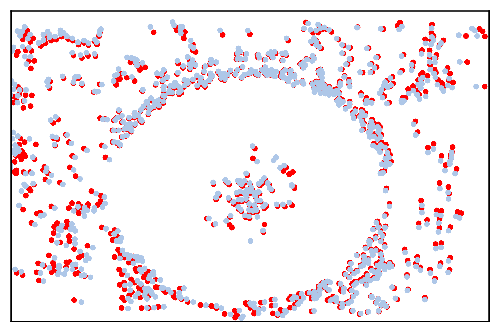}
	\caption*{Ours}
\end{subfigure}
\begin{subfigure}{.20\textwidth}
	\centering
	\includegraphics[scale=0.01,width=1.\linewidth]{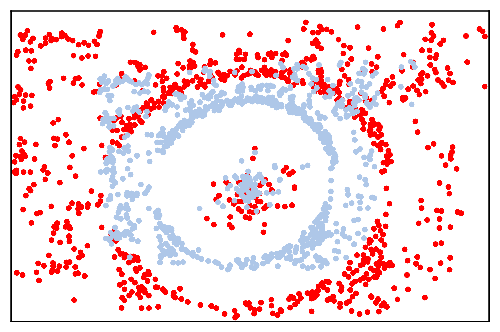}
	\caption*{OPnP}
\end{subfigure}
\begin{subfigure}{.20\textwidth}
	\centering
	\includegraphics[scale=0.01,width=1.\linewidth]{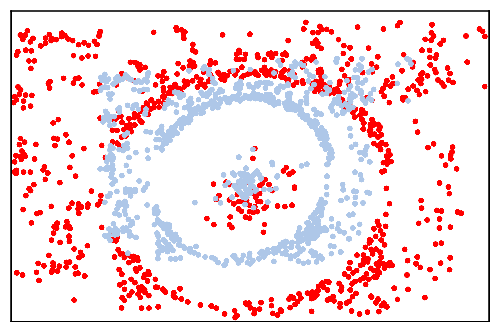}
	\caption*{EPnP+RANSAC}
\end{subfigure}
\caption{{\bf Qualitative PnP results for outlier removal.}  {\bf Left two
      columns}: Pairs of images between which we transfer the 2D points through the
    2D-3D correspondences for {\bf Top:} Florence and {\bf Bottom:} Prague.
    From the {\bf third column to the fifth column}: For each pair, we show in
    gray the reprojection of the 3D point cloud after applying the rotation and
    translation predicted by our model, OPnP, and EPnP+RANSAC,
    respectively. The red dots correspond to the ground-truth locations. Note
    that our model's predictions cover the ground truth much more closely than
    the compared methods.}
\label{fig:pnp_real}
\end{figure*}

\noindent{\bf Differentiating the eigen- and singular value decomposition.}
Whether computing the essential matrix, estimating 3D pose, or solving any other least-square problem, incorporating an eigendecomposition-solver into a deep network requires differentiating the eigendecomposition. Expressions for such derivatives have been given in~\cite{Papadopoulo00,Giles08} and reformulated in terms that are compatible with back-propagation in~\cite{Ionescu15}. Specifically, as shown in~\cite{Ionescu15}, for a matrix $\bM$ written as $\bM = \bU \bSigma \bU^\top$, the variations of the eigenvectors $\bU$ with respect to the matrix, used to compute derivatives, are
\begin{equation}
d\bU = 2\bU\left(\bK \odot (\bU^T d\bM \bU)_{sym}\right)\;,
\end{equation}
where $\bS_{sym} = \frac{1}{2}(\bS^T + \bS)$, and
\begin{equation}
\bK_{ij} = \begin{cases} \frac{1}{\sigma_i - \sigma_j}, & i\neq j \\
      0, & i=j \end{cases}\;.
      \label{eq:K}
\end{equation}
As can be seen from Eq.~\ref{eq:K}, when eigenvalues are small, 
the denominator becomes small, resulting in a large number being multiplied to the backward flow of gradients. This causes numerical instability. The effect is exaggerated when both eigenvalues of the denominator are small, even more so when they are similar $\sigma_i \approx \sigma_j$, which we found to happen in practice.
The same can be said about singular value decomposition.

A solution to this was proposed in~\cite{Papadopoulo00}, and singular- and eigen-value decomposition have been used within deep networks for problems where all the singular values are exploited and their order is irrelevant~\cite{Huang17a,Huang17b}. In the context of spectral clustering, the approach of~\cite{Law17} also proposed a solution that eliminates  the need for explicit eigendecomposition. This solution, however, was dedicated to the scenario where one seeks to use all non-zero eigenvalues, assuming a matrix of constant rank. 

Here, by contrast, we tackle problems where what matters is a single eigen- or singular-value. In this case, the order of the eigenvalues is important. However, this order can change during training, which results in a non-differentiable switch from one eigenvector to another,  as in the toy example of Section~\ref{sec:motivation}. In turn, this leads to numerical instabilities, which can prevent convergence. In~\cite{Yi18a}, we overcame this problem by first training the network using a classification loss that does not depend on eigenvectors. Only once a sufficiently good solution is found, that is, a solution close enough to the correct one for vector switching not to happen anymore, is the loss term that depends on the eigenvector associated to the smallest eigenvalue turned on. While effective, this strategy inherently relies on having access to classification labels. As such, it is ill-suited to tackle denoising scenarios. As we show later, we can achieve state-of-the-art results without the need for such a heuristic, by deriving a more robust, eigendecomposition-free loss function.

\section{Our Approach}
\label{sec:approach}

We introduce an approach to working with eigenvectors corresponding to zero eigenvalues within an end-to-end learning formalism that eliminates the gradient instabilities due to vector switching discussed in Section~\ref{sec:motivation} and the difficulties caused by repeated eigenvalues. 
To this end, we derive a loss function that directly operates on the matrix whose eigen- or singular-vectors we are interested in without explicitly performing either SVD or ED.

Below, we first discuss the generic case in which the network takes a set of of measurements as input and directly outputs the matrix of interest. We then consider two specialized but important cases. In the first,  the network outputs weights defining the matrix, which corresponds to model fitting with outlier rejection. In the second, we consider the measurements to be noisy and allow them to be adjusted to account for the noise. 
\begin{figure*}[t]
\centering
\begin{subfigure}{.25\textwidth}
	\centering
	\includegraphics[scale=0.001,width=1.\linewidth,height=4.5cm]{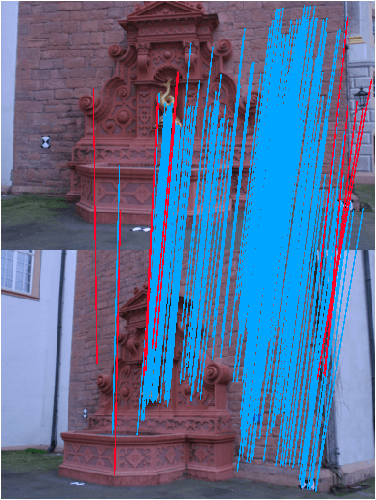}
	\label{fig:f_ours}
\end{subfigure}
\begin{subfigure}{.25\textwidth}
	\centering
	\includegraphics[scale=0.001,width=1.\linewidth,height=4.5cm]{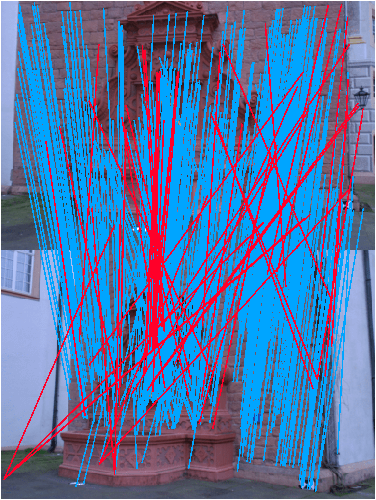}
	\label{fig:f_ransac}
\end{subfigure}
\begin{subfigure}{.223\textwidth}
	\centering
	\includegraphics[scale=0.001,width=1.\linewidth,height=4.5cm]{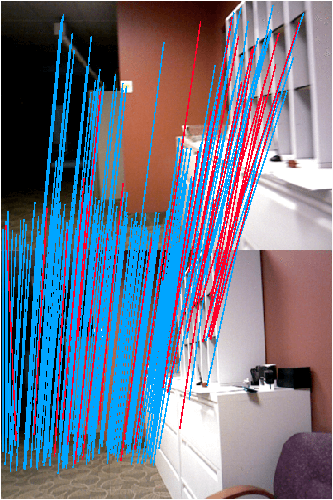}
	\label{fig:b_ransac}
\end{subfigure}
\begin{subfigure}{.223\textwidth}
	\centering
	\includegraphics[scale=0.001,width=1.\linewidth,height=4.5cm]{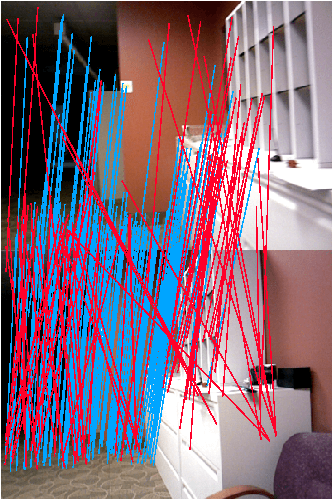}
	\label{fig:b_ours}
\end{subfigure}

\vspace{-.6em}

\centering
\begin{subfigure}{.25\textwidth}
	\centering
	\includegraphics[scale=0.001,width=1.\linewidth]{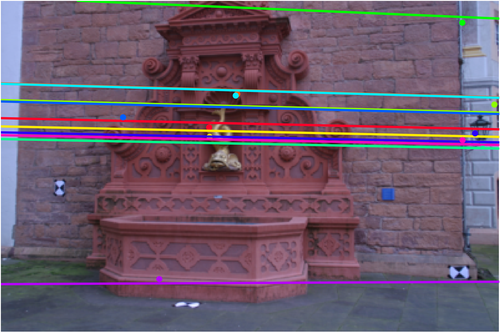}
\end{subfigure}
\begin{subfigure}{.25\textwidth}
	\centering
	\includegraphics[scale=0.001,width=1.\linewidth]{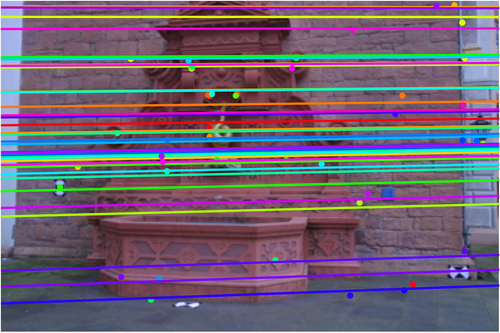}
\end{subfigure}
\begin{subfigure}{.223\textwidth}
	\centering
	\includegraphics[scale=0.001,width=1.\linewidth]{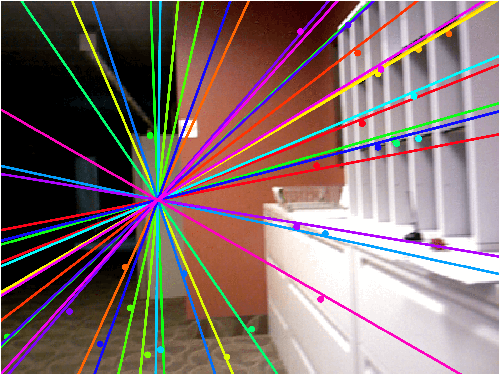}
\end{subfigure}
\begin{subfigure}{.223\textwidth}
	\centering
	\includegraphics[scale=0.001,width=1.\linewidth]{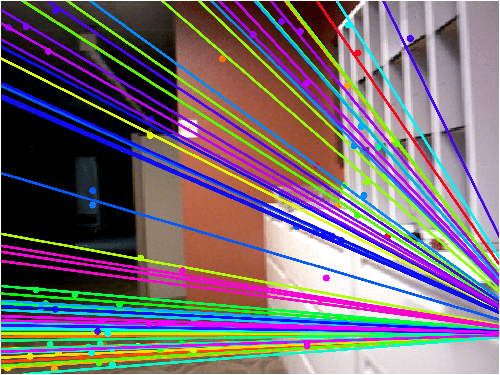}
\end{subfigure}

\centering
\begin{subfigure}{.25\textwidth}
	\centering
	\includegraphics[scale=0.001,width=1.\linewidth]{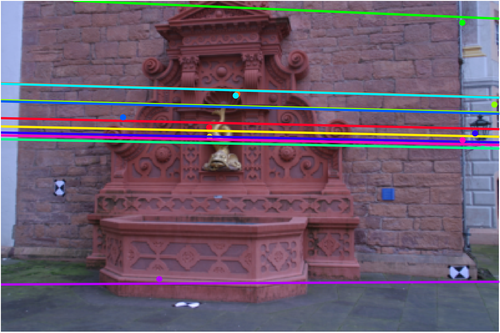}
	\caption{Ours}
\end{subfigure}
\begin{subfigure}{.25\textwidth}
	\centering
	\includegraphics[scale=0.001,width=1.\linewidth]{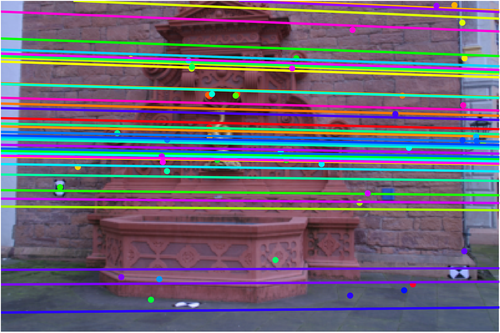}
	\caption{RANSAC}
\end{subfigure}
\begin{subfigure}{.223\textwidth}
	\centering
	\includegraphics[scale=0.001,width=1.\linewidth]{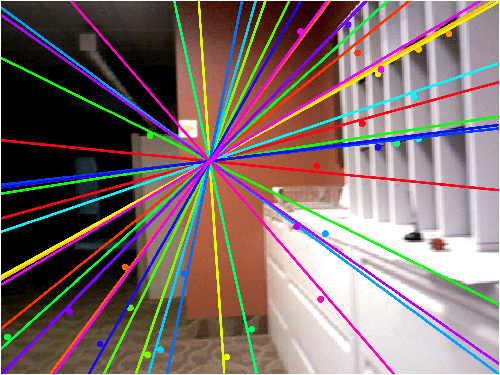}
	\caption{Ours}
\end{subfigure}
\begin{subfigure}{.223\textwidth}
	\centering
	\includegraphics[scale=0.001,width=1.\linewidth]{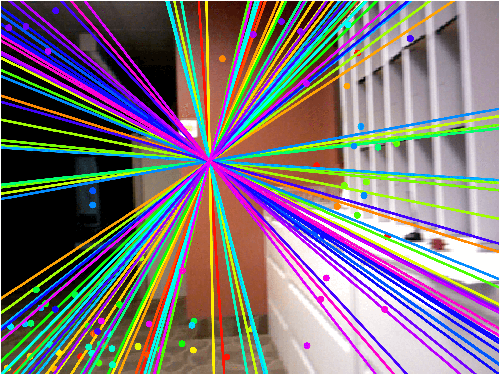}
	\caption{RANSAC}
\end{subfigure}
\caption{{\bf Qualitative comparison of our results with those of RANSAC.}
  From left to right, in each column, {\bf (a)} our results and {\bf (b)}
    RANSAC results on the ``fountain-P11'' of~\cite{Strecha08b}, {\bf (c)} our
    results and {\bf (d)} RANSAC results on the ``brown-bm-3-05'' of
    SUN3D. {\bf First two rows: } We display the correspondences that each
    algorithm labeled as inliers. The true positives are shown in blue and the
    false ones in red. The false positives of our approach are still close to
    being correct, while those of RANSAC are truly wrong. {\bf Third row:} With
    the {\it estimated} essential matrices, we plot the epipolar lines and
    their corresponding points for the false positives for each method. Note
    how, although the false positives look senseless for RANSAC, they are
    indeed following an epipolar geometry that captures both true and false
    positives. {\bf Fourth row:} Same as in the third row, but with the
    {\it ground-truth} essential matrix. Our false positives are still close to
    the true epipolar line, whereas the ones from RANSAC are not.
  }
\label{fig:key_qualitative}
\vspace{-1.5em}
\end{figure*}

\subsection{Generic Scenario}
\label{sec:general}

Given an input measurement $\bx$, let us denote by $f_{\theta}(\bx)$ the output of a deep network with parameters $\theta$. Let the network output be a matrix $\bA_\theta = f_{\theta}(\bx)$.

A seemingly natural approach to exploiting ED within the network would consist of minimizing the $\ell_2$ loss $\|\be_\theta - \bte\|^2$, where $\bte$ is the ground-truth smallest eigenvector as in~\cite{Ionescu15,Yi18a}. As discussed in Section~\ref{sec:motivation}, however, this requires differentiating the ED result to perform back-propagation and is not optimization-friendly. Furthermore, as argued in~\cite{Hartley00} for the fundamental matrix, the direct use of the $\ell_2$ norm as a distance measure is not optimal, since all the entries do not have equal importance.

\ok{To avoid the instabilities of eigenvector differentiation and use the algebraic error advocated for in~\cite{Hartley00}, we consider the standard definition of a zero eigen-value $\be_\theta$, i.e., }
\begin{equation}
\bA_\theta \be_\theta  = 0 \; \Rightarrow \; \be_\theta^\top\bA^\top_\theta\bA_\theta \be_\theta = 0 \; .
\label{eq:matrixA}
\end{equation}
This suggests defining the loss function
\begin{equation}
L_{eig}(\theta) = \bte^\top \bA^\top_\theta\bA_\theta \bte\;,
\label{eq:losseig}
\end{equation}
where $\bte$ is the ground truth zero eigen-vector. As $\bA^\top_\theta\bA_\theta$ is always positive semi-definite, $L_{eig}(\theta)$ is always greater than or equal to zero.

However, there are many matrices that satisfy Eq.~\ref{eq:losseig}, including $\bA_\theta = 0$. To rule out such trivial solutions,  we propose to minimize $L_{eig}(\theta) $ while maximizing the norm of the projection of the data vectors along the directions orthogonal to $\bte$. In the plane-fitting example of Section~\ref{sec:motivation}, this would mean maximizing the projections of the data vectors on the plane whose normal is $\bte$. The motivation behind this arises from the fact that maximizing along the directions orthogonal to $\bte$ does not preclude the correct eigenvalue to go to zero. By contrast, maximizing all eigenvalues would conflict with our original objective of Eq.~\ref{eq:losseig}, \ok{particularly because it would aim to maximize the projection of $\bA_\theta$ in the direction of $\bte$, which ideally should be zero.}

\ok{To prevent this, instead of directly regularizing $\bA_\theta$, we rely on the matrix $\bbA_\theta$ defined as}
\begin{equation}
\bbA_\theta=\bA_\theta(\bI - \bte\bte^\top) \; ,
\end{equation}
where $\bI$ is the identity matrix. Note that $\bbA_\theta\bte=\bA_\theta(\bte - \bte\bte^\top\bte)=\bA_\theta(\bte - \bte)=0$  whatever the value $\bte$. In other words, $\bbA_\theta$ projects all vectors onto the hyperplane normal to $\bte$. To prevent the trivial solutions discussed above, we would like the magnitude of these projections to be far from zero, or, in other words, the eigenvalues corresponding to directions orthogonal to $\bte$ to be as large as possible. As the trace of a matrix is the sum of its eigenvalues, this can be achieved by maximizing the trace of $\bbA^\top_{\theta} \bbA_\theta$.
We therefore introduce
\begin{equation}
L_{aux}(\theta) = -\text{tr}\left(\bbA^\top_{\theta} \bbA_\theta\right) \;
\label{eq:lossaux}
\end{equation}
as a loss to be minimized along with $L_{eig}$. 
Note that the terms defined in Eqs.~\ref{eq:losseig} and~\ref{eq:lossaux} act on the eigenvalues of the same matrix but in different directions.
$L_{eig}(\theta)$ can be rewritten as $\text{tr}\left(\bA^\top_{\theta} \bA_\theta\bte\bte^{\top}\right)$ and $L_{aux}(\theta)$ as $-\text{tr}\left(\bA^\top_{\theta} \bA_\theta(\bI - \bte\bte^{\top})\right)$.
Because of this, $L_{aux}(\theta)$ does not prevent us from obtaining a matrix with a truly zero eigenvalue in the direction $\bte$.

\begin{figure}[t]
    \centering
    \begin{subfigure}{.5\textwidth}
		\includegraphics[width=1.0\linewidth]{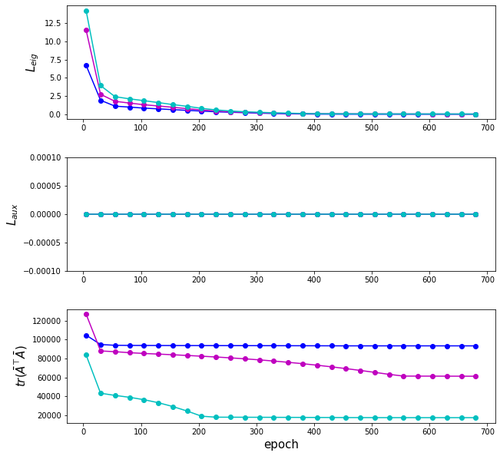}
    \end{subfigure}{}
    
 \caption{{\bf Loss evolution for different training samples.}
   \ok{This example was obtained using synthetic data for the wide-baseline stereo task. As in the plane fitting experiment, no network is involved and we directly optimize the matches weights. We set the total number of matches to be 100 and initialize all the weights to be one in the three data samples, represented as different colors: one sample with 10 outliers (blue), one with 30 outliers (magenta), and one with 50 outliers (cyan).}
   	We show, from top to bottom, our main loss term, our regularizer and the corresponding trace value $tr(\bbA^{\top} \bbA)$. In each case, we plot the evolution of the term as optimization progresses. As the weights applied to the matches remove the
       outliers, $L_{eig}$ approaches zero for all cases. Note, however, that, while
       $tr(\bbA^{\top} \bbA)$ also decreases, it remains far from zero, and more importantly, reaches final values that differ significantly for the different samples.
       By contrast, our regularizer $L_{aux}$ has much lower values that are of similar magnitudes for all samples, as indicated by the fact that the three curves overlap. This facilitates setting the weights of this regularizer.}
       
\label{fig:regularizer_laux}

\end{figure}

However, $L_{eig}(\theta)$ is bounded by zero whereas  $L_{aux}(\theta)$ is not and can become very large. %
In practice, we have observed it to reach values of ${\cal O}(10^3)$ and thus dominate the loss. \ok{Furthermore, in our deep learning context, during training, the network will produce one matrix $\bA_\theta$ for every training sample, and thus $L_{aux}$ will sum over multiple trace terms. The magnitude of these terms will then significantly vary across the training samples, putting more focus on some training samples and making it hard to define the relative weight of $L_{aux}$ and $L_{eig}$. This is illustrated by the bottom row of Figure~\ref{fig:regularizer_laux}, where we show the evolution of the trace for 3 samples. Even when all outliers are removed, the values for the 3 samples have different magnitudes.}

\ok{To overcome this and facilitate balancing the two loss terms, we use a robust kernel, $e^{-x}$, that mitigates the effects of extreme trace values and brings the regularizer values of all training sample to a commensurate magnitude in the range $[0, 1]$. The resulting training curves for the same three samples as before are shown in the middle row of Figure~\ref{fig:regularizer_laux}. Note that they are of similar magnitudes and much lower than the trace itself, thus interfering less with our main objective $L_{eig}$.
Specifically, we take our complete loss function to be}
\begin{equation}
L(\theta) = \bte^\top \bA^\top_\theta \bA_\theta \bte + \alpha\exp\left(-\beta\text{tr}\left(\bbA^\top_{\theta} \bbA_\theta\right)\right)\;,
\label{eq:general}
\end{equation}
where $\alpha$ and $\beta$ are positive scalars. \ok{Ideally, when there is a perfect solution, such that the outliers' weights are all zero and the inliers have no noise}, the value of $\alpha$ would not matter as both terms would go to zero. However, in practice, because of noise and measurement errors, this does not occur, and it is necessary to control the influence of the two terms. $\alpha$ governs whether we focus on reducing the first term -- amount of estimation error -- or the second one -- amount of information that is preserved in the non-null space according to ground truth. The former loosely relates to the accuracy of the solution, and the latter to the recall. $\beta$ controls to what degree we would like to consider the second term, as the trace value can easily go up to thousands, making it effectively reach zero without a proper $\beta$ value.
This loss is fully differentiable and can thus be used to learn the network parameters. We provide its derivatives in the next section.

\subsection{Derivatives for the General Case}

To compute the derivatives of the loss w.r.t. the parameters $\theta$, we first follow the chain rule to obtain
\begin{equation}
\begin{aligned}
\frac{\partial L(\theta)}{\partial\theta} &= \frac{\partial L_{eig}}{\partial \bA} 
\frac{\partial \bA}{\partial \theta}
+ \frac{\partial L_{aux}}{\partial \bA} 
\frac{\partial \bA}{\partial \theta}\;.
\end{aligned}
\label{eq:chain}
\end{equation}
	
We can write the first term in the equation above in index notation as
\begin{equation}
\begin{aligned}
L_{eig} &= \bte^\top \bA^\top \bA \bte
=\sum_{i=1}^{n}\sum_{j=1}^{n}\left(\sum_{k=1}^{m}\bA_{ki}\bA_{kj}\right)\bte_{i}\bte_{j}\;,
\end{aligned}
\end{equation}
where we omitted the explicit dependency of $\bA$ and $\bbA$ on $\theta$ for the sake of conciseness. Using the same notation, the second term can be written as
\begin{equation}
\begin{aligned}
L_{aux} &= \alpha\exp\left(-\beta\text{tr}\left(\bbA^\top \bbA\right)\right) \\
&=\alpha\exp\left(-\beta\sum_{i=1}^{n}\left[ \sum_{k=1}^{m}\bbA_{ki}\bbA_{kj}\right]_{ii}\right)\\
&=\alpha
\exp\left(-\beta
\sum_{i=1}^{n} \sum_{k=1}^{m}
\left( \bA_{ki} - \sum_{j=1}^{n}\bA_{kj}\bte_{j}\bte_{i} \right)^{2}\right)\;,
\end{aligned}
\end{equation}
where $\bbA = \bA(\bI - \bte\bte^\top)$, and $\bbA_{ki} = \bA_{ki} - \sum_{j=1}^{n}\bA_{kj}\bte_{j}\bte_{i}$.
Computing the derivatives with respect to $\bA_{ki}$, we obtain
\begin{equation}
\begin{aligned}
\frac{\partial L_{eig}}{\partial \bA_{ki}}
= \sum_{j=1}^{n}\bA_{kj}\bte_{i}\bte_{j} +  \sum_{i=1}^{n}\bA_{ki}\bte_{i}\bte_{j}
=\sum_{i=1}^{n}2\bA_{ki}\bte_{i}\bte_{j}
\;,
\end{aligned}
\end{equation}

\begin{equation}
\begin{aligned}
\frac{\partial L_{aux}}{\bA_{ki}} &= \alpha\left(-2\beta\left(1 - \bte_{i}\bte_{i}\right)\right)
\left( \bA_{ki} - \sum_{j=1}^{n}\bA_{kj}\bte_{j}\bte_{i} \right)\\
&\exp\left(-\beta
\sum_{i=1}^{n} \sum_{k=1}^{m}
\left( \bA_{ki} - \sum_{j=1}^{n}\bA_{kj}\bte_{j}\bte_{i} \right)^{2}\right)\\
&=-2\alpha\beta\bbA_{ki}(1 - \bte_{i}\bte_{i})\exp\left(- \beta
\sum_{i=1}^{n} \sum_{k=1}^{m} (\bbA_{ki})^{2}\right)
\;.
\end{aligned}
\end{equation}

Then, turning them back to matrix form, we write
\begin{equation}
\begin{aligned}
\frac{\partial L_{eig}}{\partial \bA} &= 2\bA\bte\bte^{\top}\;,\\
\frac{\partial L_{aux}}{\bA}&=-2\alpha\beta\bbA\text{diag}(\bI - \bte^{\top}\bte)
\exp\left(- \beta tr(\bbA^{\top}\bbA)\right)\;.
\end{aligned}
\label{eq:terms}
\end{equation}
Here, we can see that modifying $\alpha$ and $\beta$ leads to directly modifying the magnitude of the gradient. 

Finally, bringing Eq.~\ref{eq:terms} into Eq.~\ref{eq:chain} lets us write a gradient descent update of the network parameters as
\begin{equation}
\begin{aligned}
\Delta\theta &= 2\bA\bte\bte^{\top}\frac{\partial \bA}{\partial \theta} \\
&-2\alpha\beta\bbA\text{diag}(\bI - \bte^{\top}\bte)
\exp\left(- \beta tr(\bbA^{\top}\bbA)\right)
\frac{\partial \bA}{\partial \theta}\;.
\end{aligned}
\end{equation}
We then use this $\Delta\theta$ for ADAM optimization. 

Note that if we were to directly use $\bA$ instead of $\bbA$ in the second loss term, the gradients would become
\begin{equation}
\begin{aligned}
\frac{\partial L_{aux}}{\bA_{ki}} &= - (2\alpha\beta\bA_{ki})\exp\left(-\beta\sum_{i=1}^{n} \sum_{k=1}^{m}\bA_{ki}\bA_{ki}\right)\;,\\
\frac{\partial L_{aux}}{\bA}&=- (2\alpha\beta\bA)\exp\left(-\beta tr(\bA^{\top}\bA)\right)\;.
\end{aligned}
\end{equation}

In contrast to Eq.~\ref{eq:terms}, the second term contributes to the entire matrix $\bA$, including in the direction of $\bte$, which then cancels out, or at least reduces, the contribution of the $L_{eig}$ term.

\subsection{Model Fitting and Outlier Rejection}
\begin{figure*}[t]
	\centering
	\begin{subfigure}{.43\textwidth}
		\centering
		\includegraphics[scale=0.001,width=1.\linewidth]{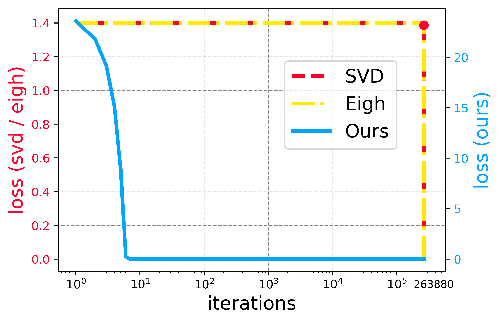}
		\caption{Loss Evolution with vanilla gradient-descent}
	\end{subfigure}
	\hspace{3em}%
	\begin{subfigure}{.43\textwidth}
		\centering
		\includegraphics[scale=0.001,width=1.\linewidth]{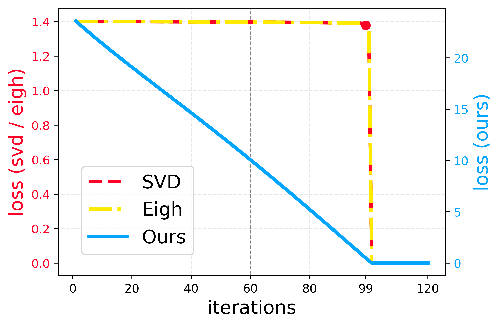}
		\caption{Loss Evolution with Adam}
	\end{subfigure}
	\caption{{\bf Plane fitting in the presence of a single outlier.}  We
          report the results for Single Value Decomposition(SVD), self-adjoint
          Eigendecomposition(Eigh), and for our loss function in
          Eq.~\ref{eq:modelFit}. {\bf(a)} Results using vanilla
            gradient-descent. To visualize all results in a single graph, we
            plot the results in log scale. {\bf (b)} Results using
            Adam~\cite{Kingma15}. Regardless of the choice of optimizers, SVD
            and Eigh slowly decrease in the beginning, followed by a sudden
            drop, denoted with the red dot. By contrast, our method converges
          gradually and smoothly}.
\label{fig:plane_fitting_single}
\end{figure*}

\begin{figure*}[t]
	\centering
	\begin{subfigure}{.32\textwidth}
		\centering
		\includegraphics[scale=0.001,width=1.\linewidth]{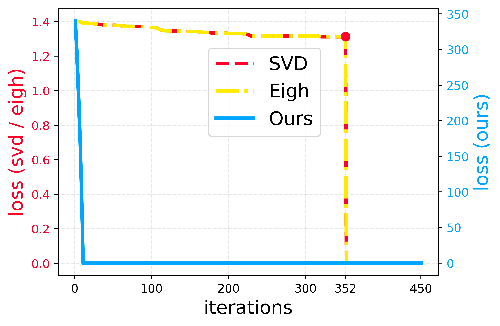}
		\caption{Loss Evolution}
		\label{fig:gd}
	\end{subfigure}
	\begin{subfigure}{.32\textwidth}
		\centering
		\includegraphics[scale=0.001,width=1.\linewidth]{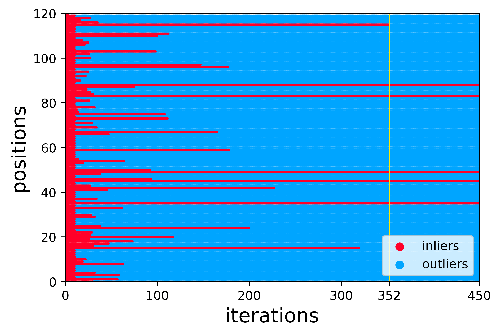}
		\caption{Inliers with SVD}
		\label{fig:in_our}
	\end{subfigure}
	\begin{subfigure}{.32\textwidth}
		\centering
		\includegraphics[scale=0.001,width=1.\linewidth]{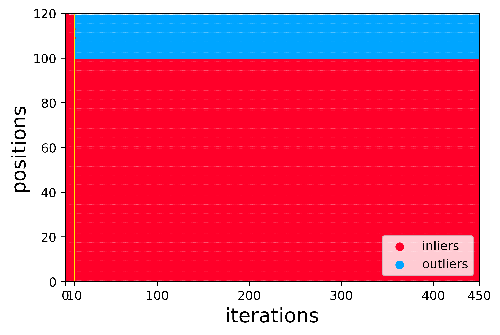}
		\caption{Inliers with Ours}
		\label{fig:in_eig}
	\end{subfigure}

	\begin{subfigure}{.62\textwidth}
		\centering
		\includegraphics[scale=0.001,width=1.\linewidth]{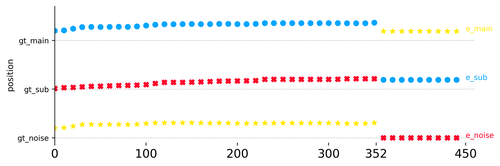}
		\caption{Switching illustration}
	\end{subfigure}
	\hspace{1em}
	\begin{subfigure}{.32\textwidth}
		\centering
		\includegraphics[scale=0.001,width=1.\linewidth]{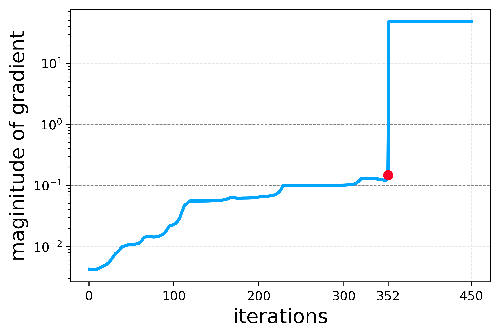}
		\caption{Gradient Evolution}
	\end{subfigure}
        \caption{{\bf Plane fitting in the presence of multiple outliers.}  We
          report results for Singular Value Decomposition (SVD), self-adjoint
          Eigendecomposition (Eigh), and our loss function.  {\bf (a)} Loss
          evolution for all methods. While our loss converges smoothly, SVD
            and Eigh remain stagnant until the shear drop at iteration 352.
            {\bf (b)} Inlier selection of SVD. The inliers chosen by SVD are shown
            in red, and the outliers in blue. Positions 1 to 100 are true
            inliers. With multiple outliers, SVD discards many inliers, as well
            as estimating outliers as inliers until iteration 352, denoted with
            the yellow vertical line, where it no longer makes mistakes. By
            contrast, as shown in {\bf (c)}, our approach correctly rejects the
            outliers and accepts all inliers, starting from iteration 10, again
            marked with the yellow vertical line. Our method also does not show
            the sporadic change in the selection of inliers visible in the case
            of SVD in {\bf (b)}. {\bf (d)} Illustration of switching. We draw
            the three eigenvectors of the ground-truth plane as the three
            horizontal lines, and the Euclidean distance to them from the
            first, second, and third eigenvectors estimated by SVD as yellow,
            blue, and red lines, respectively. Note the abrupt transition at
            iteration 352, due to the first eigenvector correctly following the
            largest ground-truth eigenvector starting from this iteration,
            whereas it was following the smallest until this iteration. This is
            clearly harmful for optimization. {\bf
              (e)} Magnitude of the gradient
            $\frac{\partial{L}}{\partial{X}}$ for each iteration with SVD,
            in log scale. Note the drastic change in iteration 352.}
            
  \label{fig:plane_fitting_multiple}
\end{figure*}

\label{sec:model}

The generic formulation presented above assumes that all measurements have the same weight and that the network predicts all the coefficients of the matrix instead of a number of design parameters from which the matrix can be computed. In practice, this is rarely the case. For problems such as plane-fitting, ellipse-fitting, PnP, and estimating the essential matrix, the data typically comes in the form of a set of $N$ measurements  $\bX_i$ for $1 \leq i \leq N$ where the $\bX_i$ are vectors of dimension $d$, where $d=3$ for plane-fitting, $d=6$ for ellipse-fitting, $d=12$ for PnP, and $d=9$ for wide-baseline stereo. In such problems, the network must be trained to assign to each measurement a weight $w_i$ between 0 and 1 such that the zero eigenvalue of the  matrix $\bX^T \bW_\theta \bX$ is the ground-truth vector $\bte$, where $\bX$ is the matrix obtained by grouping all the $\bX_i$ vectors into an $N \times d$ matrix  and $\bW_\theta$ the diagonal matrix obtained from the $N$-vector of weights $w_i$ predicted by the network.

In this scenario, the matrix $\bA_\theta$ of Eq.~\ref{eq:matrixA} becomes $\bW^{1/2}\bX$, $\bA^\top_\theta \bA_\theta$ can be written as $\bX^T \bW_\theta \bX$, and the loss function of Eq.~\ref{eq:general} becomes
\begin{equation}
L(\theta) =\bte^\top \bX^\top \bW_\theta \bX \bte\ + \alpha \exp\left(-\beta tr(\bbX^\top\bW_\theta\bbX)\right) \; .
\label{eq:modelFit}
\end{equation}

For the sake of completeness, we briefly remind the reader of how the the $\bX_i$ vectors can be obtained from image measurements  in the remainder of this section. For more details, we refer the interested readers to standard texts such as~\cite{Hartley00}. %

\subsubsection{Plane Fitting}
Let us first consider the toy outlier rejection problem used to motivate our approach in Section~\ref{sec:motivation}. Note that, for this experiment, we do not train a deep network, or perform any learning procedure. Instead, given $N$ 3D points $\bx_i$, including inliers and outliers, we directly optimize the weight $w_i$ of each point. At every optimization step, given the current weight values, we compute the weighted mean of the points $\mu = \frac{1}{\sum_{i = 1}^{N}w_{i}}\sum_{i=1}^{N}w_{i}\bx_{i}$. We then define $\bX$ to be the $3\times N$ matrix of mean-subtracted 3D points. 

\subsubsection{Ellipse Fitting}
\begin{figure*}[t]
\centering
\begin{subfigure}{.48\textwidth}
	\centering
	\includegraphics[scale=0.001,width=1.\linewidth, trim = -10 0 -10 0, clip]{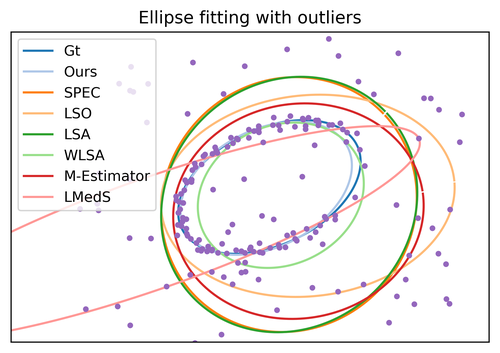}
	\caption{Outlier removal}
\end{subfigure}
\begin{subfigure}{.48\textwidth}
	\centering
	\includegraphics[scale=0.001,width=1.\linewidth, trim = -10 0 -10 0, clip]{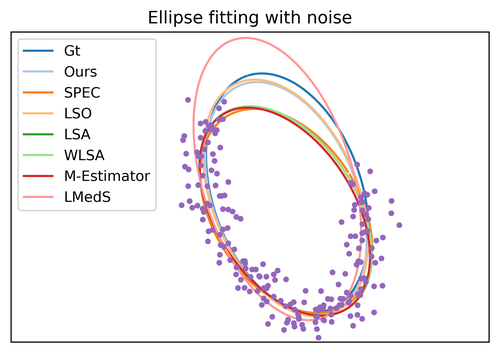}
	\caption{Denoising}
\end{subfigure}

\caption{{\bf Qualitative ellipse fitting results.} {\bf (a)}~Fitting
    results for various methods in the presence of outliers and minor
    noise. {\bf (b)}~Results in the presence of noise only. Our method gives
    the most accurate results. Note that in {\bf (a)}, our method is the only
    method that fits the true ellipse in the center well, due to the abundance
    of outliers in this setup.}
\label{fig:ellipse_fitting_qual}
\end{figure*}

\begin{figure*}[t]
\centering
\begin{subfigure}{.48\textwidth}
    \centering
    \includegraphics[scale=0.001,width=1.\linewidth, trim = -10 0 -10 0, clip]{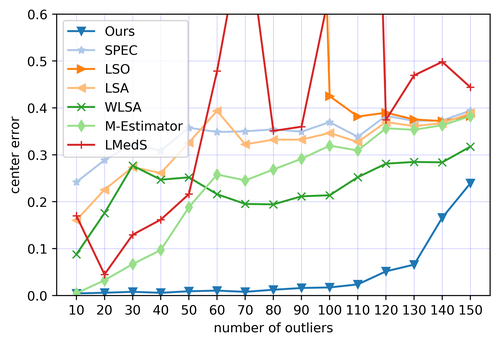}
    \caption{Outlier removal}
\end{subfigure}
\begin{subfigure}{.48\textwidth}
    \centering
    \includegraphics[scale=0.001,width=1.\linewidth, trim = -10 0 -10 0, clip]{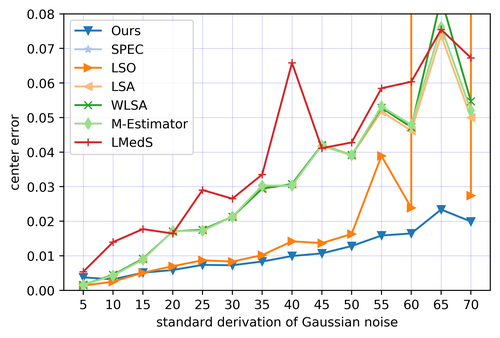}
    \caption{Denoise}
\end{subfigure}

\begin{subfigure}{.48\textwidth}
	\centering
	\includegraphics[scale=0.001,width=1.\linewidth, trim = -10 0 -10 0, clip]{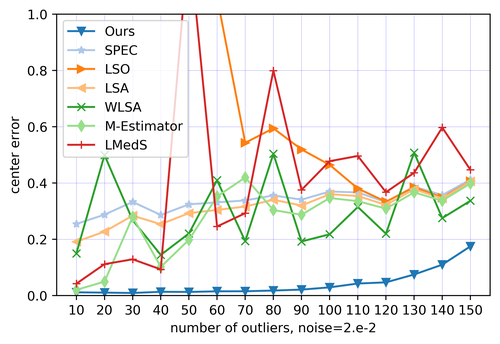}
	\caption{Outlier removal and denoising, with moderate noise}
\end{subfigure}
\begin{subfigure}{.48\textwidth}
	\centering
	\includegraphics[scale=0.001,width=1.\linewidth, trim = -10 0 -10 0, clip]{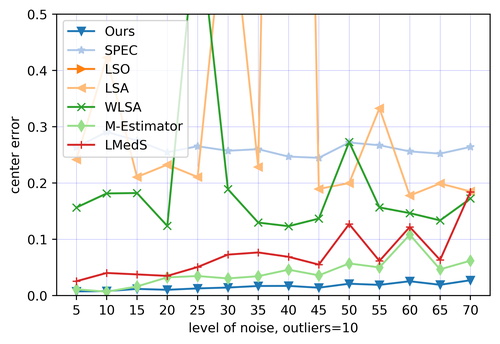}
	\caption{Outlier removal and denoising, with moderate outliers}
\end{subfigure}
\caption{{\bf Quantitative ellipse fitting results.} We report the center
    point error as in~\cite{Fitzgibbon99}. {\bf (a)} Outlier removal results. {\bf (b)}
    Denoising results. {\bf (c)} Simultaneous outlier removal and denoising
    results with moderate noise of $2 \times 10^{-2}$. {\bf (d)} Simultaneous
    outlier removal and denoising results with 10 outliers. Our method performs
    best in all cases.
  }
\label{fig:ellipse_fitting}
\end{figure*}

For this task, given a set of 2D coordinates, we aim to find the parameters of the ellipse that best fits the data. In contrast to the plane-fitting case, here we make use of a deep network, specifically the same one as in~\cite{Yi18a}. However, because the task is different than that tackled in~\cite{Yi18a}, our network takes as input $C$ 2D coordinates, not correspondences. We use this network to predict the weights $\bW$ of Eq.~\ref{eq:modelFit}, which is a $C \times C$ diagonal matrix, with one weight for each 2D coordinate.

Formally, for each 2D point $\mathbf{p}_i$, let
\begin{equation}
  \mathbf{p}_i = [x_i, y_i]^{\top}
  \;.
\end{equation}
We can then write a standard equation for an ellipse as
\begin{equation}
Ax^{2} + 2Bxy + Cy^{2} + 2Dx+ 2Ey + F = 0
\;\;,
\label{eq:ellipse_standard_form}
\end{equation}
where $A$, $B$, $C$, $D$, $E$, and $F$ define the parameters of the ellipse. We can therefore write
\begin{equation}
  \bX^{(i)} = [x_i^2, 2x_iy_i, y_i^2, 2x_i, 2y_i, 1]\;,
  \label{eq:datamat_ellipse}
\end{equation}
for this ellipse fitting case.

\subsubsection{Perspective-n-Points}

The goal of this problem is to determine the absolute pose (rotation and translation) of a calibrated camera, given known 3D points and corresponding 2D image points. For this task, we again use a deep network with the same architecture as that in~\cite{Yi18a}. Here, however, the network takes as input $C$ correspondences between 3D and 2D points and outputs a $C$-dimensional vector of weights, one for each correspondence.

Mathematically, we can denote the input correspondences as
\begin{equation}
\mathbf{q}_{i} = [x_{i}, y_{i}, z_{i}, u_{i}, v_{i}]^\top\;,
\end{equation}
where $x_{i}, y_{i}, z_{i}$ are the coordinates of a 3D point, and $u_i$, $v_i$ denote the corresponding image location.
According to~\cite{Hartley00}, we have
\begin{equation}
f_{\text{scale}}\begin{bmatrix}
u_{i}\\
v_{i}\\
1
\end{bmatrix}\!{=}\!
\begin{bmatrix}
\mathbf{R}{,}\mathbf{t}
\end{bmatrix}
\begin{bmatrix}
x_{i}\\
y_{i}\\
z_{i}\\
1
\end{bmatrix}\!{=}\!
\begin{bmatrix}
 p_{1} &p_{2} &p_{3} &p_{4} \\ 
 p_{5} &p_{6} &p_{7} &p_{8} \\ 
 p_{9} &p_{10} &p_{11} &p_{12} \\
\end{bmatrix}
\begin{bmatrix}
x_{i}\\
y_{i}\\
z_{i}\\
1
\end{bmatrix}\;.
\end{equation}
Following the Direct Linear Transform (DLT) method~\cite{Hartley00}, the transformation parameters can be obtained as the zero-valued singular vector of a data matrix $\bX \in \mathbb{R}^{2C\times12}$, every two rows of which are computed from one correspondence $\bq_{i}$ as
\setcounter{MaxMatrixCols}{12}
\small
\begin{equation}
\begin{bmatrix}
\mathbf X^{(2i{-}1)}\\
\mathbf X^{(2i)}
\end{bmatrix}\! {=} \! 
\begin{bmatrix}
 x_{i},  y_{i}, z_{i}, 1, \; 0, \; 0, \; 0, \; 0,  -u_{i}x_{i},  -u_{i}y_{i},  -u_{i}z_{i},  -u_{i} \\
 0, \; 0, \; 0, \; 0, \; x_{i}, y_{i}, z_{i}, 1,  -v_{i}x_{i},  -v_{i}y_{i},  -v_{i}z_{i},  -v_{i}
\end{bmatrix}
\;,
\label{eq:datamat_pnp}
\end{equation} \normalsize
where $\bX^{(i)}$ denotes row $i$ of $\bX$. We can then directly use this data matrix in Eq.~\ref{eq:modelFit}. As suggested by~\cite{Hartley00} and also done in~\cite{Yi18a}, we use the normalized coordinate system for the 2D coordinates.
We assume that the correspondences for this setup are given, as in~\cite{Yi18a}

Note that the characteristics of the rotation matrix, that is, orthogonality and determinant 1, are not preserved by the DLT solution. Therefore, to make the result a valid rotation matrix, we refine the DLT results by the generalized Procrustes algorithm~\cite{Garro12,Schonemann66}, which is a common post-processing technique for PnP algorithms. Note that this step is not involved during training, but only in the validation process to select the best model and at test time.

\subsubsection{Essential Matrix Estimation}

This task is the same as that studied in~\cite{Yi18a}, that is, estimating camera motion from 2D-to-2D correspondences.
A critical downside of~\cite{Yi18a} is the use of the DLT to regress to the camera pose, which is solved via SVD/ED.
  This method therefore suffers from the switching problem discussed in Section~\ref{sec:motivation}.
  By contrast, we convert the DLT problem to a cost that directly reflects the original task.

To isolate the effect of the loss function only, we follow the same setup as in~\cite{Yi18a}. Specifically, we use the same network architecture as in~\cite{Yi18a}, which takes $C$ correspondences between two 2D points as input and outputs a $C$-dimensional vector of weights, that is, one weight for each correspondence.
For the same purpose, we also use the same strategy for creating correspondences---a simple nearest neighbor matching from one image to the other.

Formally, let
\begin{equation}
\mathbf{q}_{i} = \left[ u_{i}, v_{i}, u'_{i}, v'_{i}\right]^\top\;,
\end{equation}
encode the coordinates of correspondence $i$ in the two images. Following the 8-point algorithm~\cite{Longuet-Higgins81}, we construct a matrix $\bX \in \mathbb{R}^{C\times9}$, each row of which is computed from one correspondence vector $\bq_i$ as
\begin{equation}
\bX^{(i)} = [u_iu_i', u_iv_i, u_i, v_iu_i', v_iv_i', v_i, u_i', v_i', 1]\;,
\end{equation}
where $\bX^{(i)}$ denotes row $i$ of $\bX$. We can then directly use this matrix in the loss of Eq.~\ref{eq:modelFit}, which corresponds to a weighted version of the 8-point algorithm~\cite{Zhang98}.

Note that, as suggested by~\cite{Hartley00} and done in~\cite{Yi18a}, we normalize the 2D coordinates to $[-1,1]$ using the camera intrinsics as input to the network. Furthermore, when calculating the loss, as suggested by~\cite{Hartley97}, we move the centroid of the reference points to the origin of the coordinate system and scale the points so that their RMS distance to the origin is equal to $\sqrt{2}$. This means that we also have to scale and translate $\bte$ accordingly.

Here, our formulation originates from the DLT. As a result, our loss minimizes the algebraic distance, which could be suboptimal. While there exist other formulations for solving for the essential matrix, such as the gold-standard method that minimizes the geometric distance~\cite{Hartley00}, these methods rely on iterative minimization. Including such an iterative process in a loss function can be achieved by solving a bi-level optimization problem, as discussed in~\cite{Gould16}. However, this requires second-order derivatives, and thus may become expensive in the context of deep networks, due to the large number of parameters. Alternatively, one can unroll a fixed number of iterations, but this number needs to be determined and this requires an initialization. Our method can be thought of as a way to provide such an initialization. We leave the task of incorporating a nonlinear refinement in the training process given our initialization as future work, and focus on the DLT case, which allows us to demonstrate the benefits of our approach over direct eigenvalue minimization. \ok{Note that, as an alternative, other loss functions can be used, such as a classification loss on the keypoint weights~\cite{Yi18a} or the residual error using the symmetric epipolar distance~\cite{Ranftl18}. These methods, however, require defining a threshold, whose precise value can have a significant impact on the results.}

\subsection{Handling Noisy Measurements}
\label{sec:denoise}

Beside outliers, real-world measurements are typically affected by noise, which
deteriorates the quality of the estimate of interest.  Here, we show that our
framework also allows us to address denoising problems in addition to
  outlier rejection, by making the network estimate a refinement of the input
measurements.  In other words, while outlier rejection can be expressed by
estimating a multiplicative factor, that is $\bW$, denoising translates to
predicting an additive one.

To remove noise from input data, we propose to train our networks to
  output, in addition to the weights $\bW$, a denoising matrix $\Delta\bX$ that
  has the same dimension as the input data. Thus, instead of directly using
  $\bX$, we create the data matrix with the denoised data, $\bXe = \bX + \Delta\bX$. However,
  directly using the denoised data matrix $\bXe$ instead of $\bX$ in
  Eq.~\ref{eq:modelFit} is problematic as, with the denoising vector, one could add arbitrary
  displacements in the direction corresponding to the largest eigenvalues to
  reach very large values. 
  \ok{To overcome this, we rely on the assumption that the displacements should be small, which can be encoded by the constraint $\|\Delta x_i\| \leq c$ for a given threshold $c$ corresponding to the noise value. To avoid using hard constraints, which are difficult to handle with deep networks, we use soft constraints, which can be thought as the Lagrangian version of the hard ones.}
  Altogether, we thus define our loss as
  \begin{equation}
    \begin{aligned}
      L(\Delta{\bx}, W) =
      &\bte^\top \bXe^\top\bW \bXe \bte + 
      \alpha\exp\left(-\beta\text{tr}\left(\bbX^\top\bW \bbX\right)\right) +\\
      &\gamma\frac{1}{N}\sum_{i=1}^{N}\left\Vert\Delta{\bx_{i}}\right\Vert ,
    \end{aligned}
    \label{eq:ell_comb}
  \end{equation}
  where $\bte$ is the ground-truth eigenvector,
  $\Delta{\bx_{i}} = [\Delta{x_{i}}, \Delta{y_{i}}]$ and
  $\bbX {=} \bX(\bI - \bte\bte^\top)$, expressed in terms of the original
  measurements, not the denoised ones. We found this to be more stable than using $\bXe$, because the latter would still favor generating large displacements along the directions orthogonal to $\bte$, which contradicts the last loss term and deteriorates the distinction between inliers and outliers, as outliers can then be handled using either a zero weight $w$ or a large displacement $\Delta{\bx}$.

We apply the denoising term for the task of ellipse fitting and PnP. For
  both cases, we only correct for the noise in the 2D data, that is, $x_i$ and
  $y_i$ in Eq.~\ref{eq:datamat_ellipse} and $u_i$ and $v_i$ in Eq.~\ref{eq:datamat_pnp}.

We exclude essential matrix estimation from this setup, for the following reasons. First, for
  wide-baseline stereo, the same level of noise can have drastically different
  effects on the estimation, depending on the distance to the epipoles. For
  example, even the slightest noise near an epipole can greatly harm
  estimation. Second, and more importantly, the outlier ratio in wide-baseline stereo is much larger than in other tasks, e.g., 91\% in the cogsci2-05 sequence from the SUN3D dataset. As such, the positive and negative labels are highly unbalanced, and the negative labels  dominate the denoising operation, which translates to the positive ones being removed. As a consequence performing outlier removal alone has proven more robust than jointly with denoising, particularly because our network filters the matches so that the inliers attain a large majority, and the final RANSAC step then further increases robustness.

\section{Experiments}
\begin{figure*}[t]
\centering
\begin{subfigure}{.48\textwidth}
    \centering
    \includegraphics[scale=0.001,width=1.\linewidth, trim = -10 0 -10 0, clip]{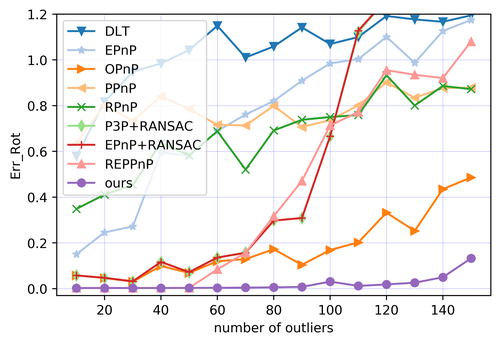}
    \caption{Rotation error}
\end{subfigure}
\begin{subfigure}{.48\textwidth}
    \centering
    \includegraphics[scale=0.001,width=1.\linewidth, trim = -10 0 -10 0, clip]{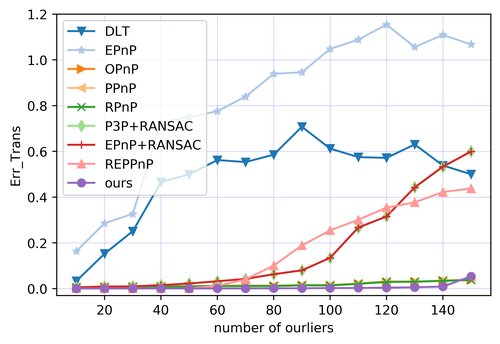}
    \caption{Translation error}
\end{subfigure}
\caption{{\bf Quantitative PnP results for outlier removal.} {\bf (a)}
    Rotation and {\bf (b)} translation errors for our method and several
  baselines. Our method gives extremely stable results despite the abundance of
  outliers, whereas all compared methods perform significantly worse as the
  number of outliers increases. Even when these methods perform well on either
  rotation or translation, they do not perform well on both. By contrast, Ours
  yields near zero errors for both measures up to 130 outliers (i.e., 65\%). }
\label{fig:pnp_outlier_removing}
\end{figure*}

\begin{figure*}[t]
\centering
\begin{subfigure}{.48\textwidth}
    \centering
    \includegraphics[scale=0.001,width=1.\linewidth, trim = -10 0 -10 0, clip]{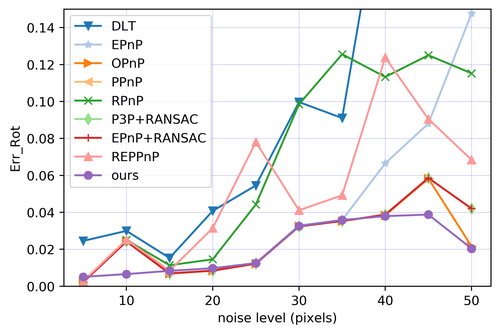}
    \caption{Rotation error}
\end{subfigure}
\begin{subfigure}{.48\textwidth}
    \centering
    \includegraphics[scale=0.001,width=1.\linewidth, trim = -10 0 -10 0, clip]{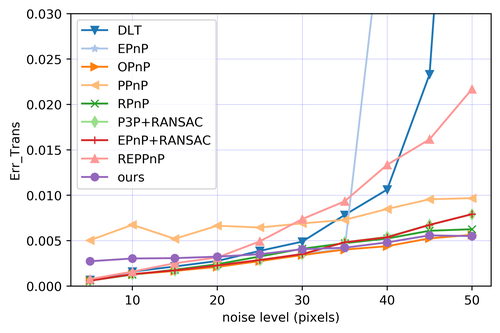}
    \caption{Translation error}
\end{subfigure}
\caption{{\bf Quantitative PnP results for denoising.} {\bf (a)} Rotation
    and {\bf (b)} translation errors for our method and several baselines with
    varying amount of additive noise on the 2D coordinates. Our method gives
    the best results in terms of rotation, and performs similarly to the best
    performing method, OPnP, for translation. Note that for translation, most
    methods, including ours, give extremely low error.}
\label{fig:pnp_denoise}
\end{figure*}

\begin{figure*}[t]
\centering
\begin{subfigure}{.48\textwidth}
    \centering
    \includegraphics[scale=0.001,width=1.\linewidth, trim = -10 0 -10 0, clip]{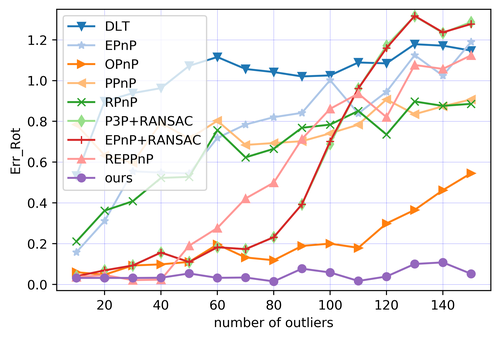}
    \caption{Rotation error}
\end{subfigure}
\begin{subfigure}{.48\textwidth}
    \centering
    \includegraphics[scale=0.001,width=1.\linewidth, trim = -10 0 -10 0, clip]{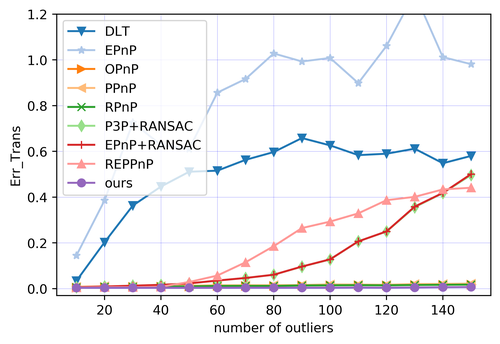}
    \caption{Translation error}
\end{subfigure}

\begin{subfigure}{.48\textwidth}
	\centering
	\includegraphics[scale=0.001,width=1.\linewidth, trim = -10 0 -10 0, clip]{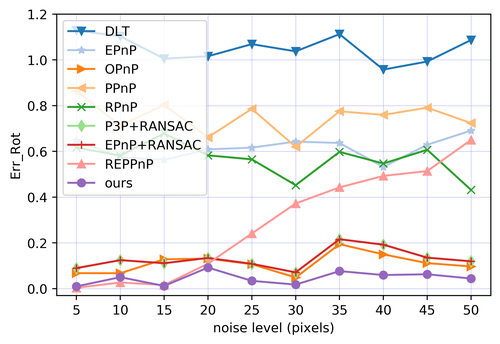}
	\caption{Rotation error}
\end{subfigure}
\begin{subfigure}{.48\textwidth}
	\centering
	\includegraphics[scale=0.001,width=1.\linewidth, trim = -10 0 -10 0, clip]{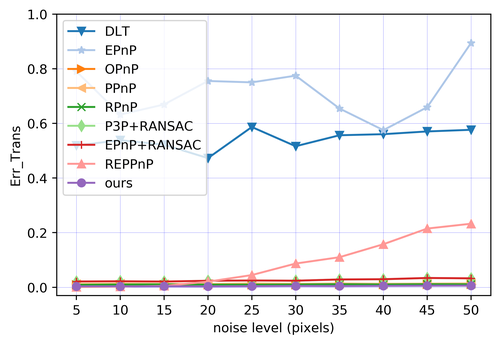}
	\caption{Translation error}
\end{subfigure}
\caption{{\bf Quantitative PnP results for simultaneous outlier removal and
      denoising.} {\bf (a)} Rotation and {\bf (b)} translation errors for
    varying number of outliers with additive Gaussian noise with standard
    deviation of 20 pixels on the 2D coordinates.  {\bf (c)} Rotation and {\bf
      (d)} translation errors for varying amount of additive noise on the 2D
    coordinates with 50 outliers. Our method gives the best results in all
    cases.
  }
\label{fig:pnp_combination}
\end{figure*}

\begin{figure*}[t]
	\centering
	\begin{subfigure}{.48\textwidth}
		\centering
		\includegraphics[scale=0.001,width=1.\linewidth]{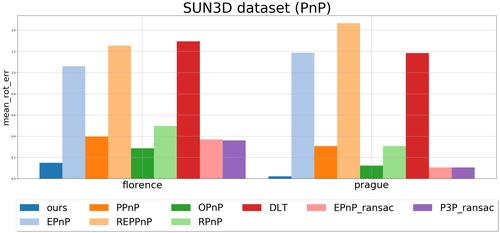}
		\caption{Rotation error}
	\end{subfigure}
	\begin{subfigure}{.48\textwidth}
		\centering
		\includegraphics[scale=0.001,width=1.\linewidth]{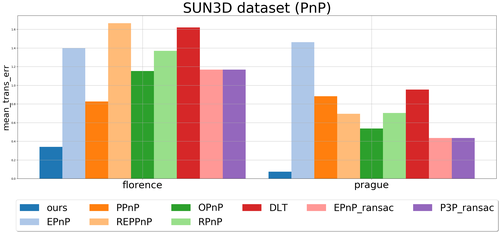}
		\caption{Translation error}
	\end{subfigure}
	\caption{\ok{{\bf Quantitative PnP results on real scenes.} The results are averaged over 1000 images. Note that our method gives the best results.}}.
\label{fig:pnp_real_all_seq}
\end{figure*}

We now present our results for the four tasks discussed above, that is, plane fitting as in Section~\ref{sec:motivation}, ellipse fitting, solving the PnP problem, and distinguishing good keypoint correspondences from bad ones.
We rely on a  TensorFlow implementation using the Adam~\cite{Kingma15} optimizer, with a learning rate of $10^{-4}$, unless stated otherwise, and default parameters.
When training a network, for ellipse fitting, keypoint matching and PnP, we used mini-batches of 32 samples. We used the same network as in~\cite{Yi18a} for all three tasks. The network takes $C$ observations as input, which can be either 2D locations for ellipse fitting, 3D-to-2D correspondences for PnP, or 2D-to-2D correspondences for wide-baseline stereo, and produces a $C$-dimensional array of weights, one for each observation. The weights indicate the likelihood of each observation to be an inlier. The observations are processed independently by weight-sharing perceptrons. The network consists of 12 residual blocks. In the each residual block, Context Normalization is applied before batch normalization and ReLU to aggregate the global information. The code to reproduce our experiments is available on Github\footnote{\href{https://github.com/Dangzheng/Eig-Free-release}{https://github.com/Dangzheng/Eig-Free-release}}.

\subsection{Plane Fitting}
\label{sec:planefitting}

We first evaluate the effectiveness of the proposed method in a simple toy setup, where the task is to remove outliers.
The setup is the one discussed in Section~\ref{sec:motivation}. We randomly sampled 100 3D points on the $z=1$ plane. Specifically, we uniformly sampled $x\in[0,40]$ and $y\in[0,2]$. We then added zero-mean Gaussian noise with standard deviation $0.001$ in the $z$ dimension. We also generated outliers in a similar way, where $x$ and $y$ are uniformly sampled in the same range, and $z$ is sampled from a Gaussian distribution with mean 50 and standard deviation 1. For the baselines that directly use the analytical gradients of SVD and ED, we take the objective function to be $\min\left\|\be_{min}(\bw) \pm \be_{gt}\right\|_{2}$, where $\be_{min}(\bw)$ is the minimum eigenvector of $\bX^\top\bW\bX$ and $\be_{gt}$ is the ground-truth noise direction ($\bte$ in Eq.~\ref{eq:modelFit}), which is also the plane normal and is the vector $[0, 0, 1]$ in this case. Note that we consider both $+\be_{gt}$ and $-\be_{gt}$ take the minimum distance, denoted by the $\pm$ and the $\min$ in the loss function. For this problem, both solutions are correct due to the sign ambiguity of ED, which should be taken into account.

We consider two ways of computing analytical gradients, one using the SVD and the other the self-adjoint eigendecomposition (Eigh), which both yield mathematically valid solutions.
To implement our approach, we rely on Eq.~\ref{eq:modelFit}, as we are mostly interested in removing outliers for this experiment.
For the learning rate,  we use $10^{-2}$ for the single outlier case, and $10^{-1}$ for the multiple outlier one.

Fig.~\ref{fig:plane_fitting_single} shows the evolution of the loss as the optimization proceeds when using Adam and vanilla gradient descent with a single outlier.
Note that SVD and Eigh have exactly the same behavior because they constitute two equivalent ways of solving the same problem.
Using Adam in conjunction with either one initially yields a very slow decrease in the loss function, until it suddenly drops to zero after hundreds of iterations. In case of vanilla gradient descent, SVD and Eigh require hundreds of thousands of optimization steps to converge. By contrast, our approach produces a much more gradual decrease in the loss.

We further show in Fig.~\ref{fig:plane_fitting_multiple} how our formulation works compared to that of SVD- and Eigh-based methods in the presence of twenty outliers.
As the problem is harder, SVD- and Eigh-based losses take more iterations to converge than in the single outlier case.
With Adam, the convergence suddenly happens after 352 iterations, when a switch of the eigenvector with the smallest eigenvalue occurs, as shown in Fig.~\ref{fig:plane_fitting_multiple} (d).

In Fig.~\ref{fig:plane_fitting_multiple} (b) and (c), we show the inliers detected during optimization. In a plane fitting example, only three inliers are required to solve the problem, thus not all inliers are used as shown by Fig.~\ref{fig:plane_fitting_multiple} (b).
However, as our formulation encourages the inclusion of inliers in the optimization through the second term in Eq.~\eqref{eq:modelFit}, we are able to recover {\em all} inliers as shown in Fig.~\ref{fig:plane_fitting_multiple} (c).

In Fig.\ref{fig:plane_fitting_multiple} (d) and (e), we can clearly see the abrupt change in the optimization at iteration 352. In Fig.\ref{fig:plane_fitting_multiple} (d), we plot the distance between the ground-truth and the estimated eigenvectors obtained through eigendecomposition, where the zero-point for each eigenvector is placed differently for better visualization. To be specific, we place the eigen-vector to the ground-truth position which closest to it. The distance between the point and the ground-truth position which below it indicates how close they are. Also for better visualization, we set the interval to be 10 and down sample the points. One can clearly see that until iteration 352, the eigenvectors are wrongly assigned, which is then suddenly corrected. In Fig.~\ref{fig:plane_fitting_multiple} (e), we show the devastating effect this switch causes; the magnitude of the gradient increases drastically, which is potentially harmful for optimization. Specifically, at this point, most of the inliers have been removed by SVD and Eigh, and thus the smallest two eigenvalues are close to zero. Therefore, because of Eq.~\ref{eq:K}, the gradients remain very large, even after this switch happens.

While in this simple plane fitting example the SVD- or Eigh-based methods converge, in more complex cases, this is not always true, as we will show next with different applications.

\subsection{Ellipse Fitting}

To evaluate our approach on the task of ellipse fitting, we built a synthetic dataset.
We generated ten thousand ellipses with random centers in the rectangle whose top right corner coordinate is $[0.5, 0.5]$ and bottom left corner coordinate is $[-0.5 ,-0.5]$, angles in the range $[0, \pi]$, long and short axes in the range $[0.2, 1]$, thus their top, right, bottom, and left coordinates are 1, 1, -1, and -1, respectively.
Working in this confined range allows us to create ellipses that are distinguishable from noise and outliers with the human eye, and yet have enough variability to capture diverse types of ellipses.
While this setup is constrained to a given scale range, in practice, ellipses at different scale ranges can
easily be dealt with by normalizing and denormalizing as in~\cite{Hartley97}.

We compare our method against other ellipse fitting techniques that are discussed in~\cite{Zhang95c}.
In particular, we evaluate the robust version of the ellipse-Specific method (Spec), Least-Squares based on Orthogonal distance (LSO), Least-Squares based on Algebraic distance (LSA), gradient Weighted Least-Squares based on Algebraic distance (WLSA), ellipse fitting with M-Estimator using the Cauchy loss function, and ellipse fitting with Least Median of Squares (LMedS).
As in~\cite{Fitzgibbon99}, we report the error of the estimated center point with arbitrary orientation for each method.

\subsubsection{Outlier Removal}

For this experiment, we incorporate outliers to our dataset by adding random points within the unit rectangle around the origin. Specifically, we randomly select $[0, 100]$ of the 200 2D points and convert them to outliers. We further add a Gaussian noise with standard deviation of $1\times 10^{-2}$.
We set $\alpha=1$ and $\beta=5\times 10^{-3}$ and rely on \Eq{modelFit}, as we would like to evaluate outlier removal only.

Quantitative results are summarized in Fig.~\ref{fig:ellipse_fitting} (a), and qualitative results are shown in Fig.~\ref{fig:ellipse_fitting_qual} (a). Our method outperforms all other methods. Note that, in Fig.~\ref{fig:ellipse_fitting_qual} (a), our method is the only one able to recover an ellipse that fits the ground-truth points.

\subsubsection{Denoising}

To test the performance of our method on denoising, we add Gaussian noise to the points, but do not include any outliers. We vary the noise level from zero to $5\times 10^{-2}$. Note that, in this pure denoising scenario, we do not predict a weight matrix $\bW$, and the outputs of the network are the displacements $\Delta{\bx_{i}}$ only. Since we penalize overly large displacements $\Delta{\bx_{i}}$, our formulation doesn't suffer from the trivial solution where the matrix output by the network goes to zero. There is no need to encourage the projections in the direction orthogonal to the eigenvector of interest to remain large, so we set $\alpha=0$ and $\beta=0$ in Eq.~\ref{eq:ell_comb}.

As in the outlier rejection case, we report the center point error. We run each experiment 100 times and report the average. The results are summarized in Fig.~\ref{fig:ellipse_fitting} (b). We also show a qualitative example in Fig.~\ref{fig:ellipse_fitting_qual} (b). Our method gives the most robust results with respect to the additive Gaussian noise.

\subsubsection{Simultaneous Outlier Removal and Denoising}

Finally, we combine both sources of perturbations and randomly choose up to 150 outliers among the 200 points and add a Gaussian noise with a standard deviation up to $0.05$.
For the loss function in Eq.~\ref{eq:ell_comb}, we set $\alpha=1$, $\beta=5\times 10^{-3}$ and $\gamma=10^{-2}$.
	
During testing, we fix either the number of outliers or the degree of noise and vary the other.
Specifically, to show robustness against noise in the presence of outliers, we fix the number of outliers to 10 and vary the standard derivation of Gaussian noise in the range $[5\times 10^{-3}, 7\times 10^{-2}]$.
To show robustness to outliers, we fix the Gaussian noise to be $2\times 10^{-2}$ and vary the number of outliers in the range $[0, 150]$. 
The results are summarized in Fig.~\ref{fig:ellipse_fitting} (c) and (d), where we can clearly see that we outperform the baselines.

\subsection{Perspective-n-Points}
Let us now turn to the PnP problem. For this application, as discussed below, we perform either outlier removal, or denoising, or both tasks jointly.

\subsubsection{Outlier Removal on Synthetic Data}
\label{sec:pnp_outlier}

Following standard practice for evaluating PnP algorithms~\cite{Lepetit09,Ferraz14}, we generate a synthetic dataset composed of 3D-to-2D correspondences with noise and outliers. Each training example comprises two hundred 3D points, and we set the ground-truth translation of the camera pose $\bt_{gt}$ to be their centroid.
We then create a random ground-truth rotation $\bR_{gt}$, and project the 3D points to the image plane of our virtual camera. As in REPPnP~\cite{Ferraz14}, we apply Gaussian noise with a standard deviation of $5$ to these projections. We generate random outliers by assigning 3D points to arbitrary valid 2D image positions. 

To focus on outlier removal only, we again first use the formulation in Eq.~\ref{eq:modelFit}. For the hyper-parameters in Eq.~\ref{eq:modelFit}, we empirically found that $\alpha=10$ and $\beta = 5\times10^{-3}$ works well for this task. During training, we randomly select between 0 and 150 of the two hundred matches and turn them into outliers.  In other words, the two hundred training matches will contain a random number of outliers that our network will learn to filter out.

We compare our method against modern PnP methods, EPnP~\cite{Lepetit09}, OPnP~\cite{Zheng13}, PPnP~\cite{Garro12}, RPnP~\cite{Li12c} and REPPnP~\cite{Ferraz14}. We also evaluate the DLT~\cite{Hartley00}, since our loss formulation is based on it. Among these methods, REPPnP is the one most specifically designed to handle outliers. As in the keypoint matching case, we tried to compute the results of a network relying explicitly on eigendecomposition and minimizing the $\ell_2$ norm of the difference between the ground-truth eigenvector and the predicted one. However, we found that such a network was unable to converge.
We also report the performance of two commonly used baselines that leverage RANSAC~\cite{Fischler81}, P3P~\cite{Kneip11}+RANSAC and EPnP+RANSAC. %
For the other methods, RANSAC didn't bring noticeable improvements, so we omitted them in the graph for better visual clarity. For all baselines, we used the default parameters provided with their implementation, which were tuned on equivalent synthetic datasets. For the RANSAC-based ones, we used the default RANSAC parameter from OpenCV. Note that our method could also make use of RANSAC, but we use it on its own here. While tuning the RANSAC parameter would affect the final results, all the methods, including ours if we were to use RANSAC, would benefit from it, which would leave our conclusions unchanged.

For this comparison, we use standard rotation and translation error metrics~\cite{Crivellaro18}. 
To demonstrate the effect of outliers at test time, we fix the number of matches to 200 and vary the number of outliers from $10$ to $150$. We run each experiment 100 times and report the average.

Fig.~\ref{fig:pnp_outlier_removing} summarizes the results. We outperform all other methods significantly, especially when the number of outliers increases. REPPnP is the one competing method that seems least affected by the outliers. As long as the number of outliers is small, it is on a par with us. However, passed a certain point---when there are more than 40 outliers, that is, 20\% of the total number of correspondences---its performance, particularly in terms of rotation error, decreases quickly, whereas ours does not.

\subsubsection{Outlier Removal on Real Data}

We then evaluate our PnP outlier removal approach on the real dataset of~\cite{Heinly15}. Specifically, the 3D points in this dataset were obtained using the SfM algorithm of~\cite{Wu13}, which also provides a rotation matrix and translation vector for each image. We treat these rotations and translations as ground truth to compare different PnP algorithms. Given a pair of images, we extract SIFT features at the reprojection of the 3D points in one image, and match these features to SIFT keypoints detected in the other image. This procedure produces erroneous correspondences, which a robust PnP algorithm should discard. In this example, we used the model trained on the synthetic data described before. Note that we apply the model {\bf without any fine-tuning}, that is, the model is only trained with purely synthetic data. We observed that, except for EPnP+RANSAC, OPnP and P3P+RANSAC, the predictions of the baselines are far from the ground truth, which led to points reprojecting outside the image. 

In Fig~\ref{fig:pnp_real_all_seq}, we provide quantitative results on this real dataset. These results are averaged over 1000 images. In Fig.~\ref{fig:pnp_real}, we compare the reprojection of the 3D points in the input image after applying the rotation and translation obtained with our model and with OPnP and EPnP+RANSAC. Note our better accuracy, both quantitatively and qualitatively.

\begin{figure*}[htb]
\centering
\begin{subfigure}{0.48\textwidth}
	\centering
	\includegraphics[scale=0.001,width=1.\linewidth]{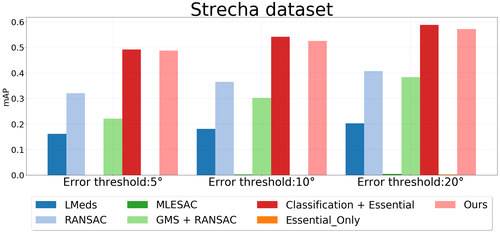}
	\label{fig:arc_strecha}
\end{subfigure}
\hspace{+.5em}
\begin{subfigure}{0.48\textwidth}
	\centering
	\includegraphics[scale=0.001,width=1.\linewidth]{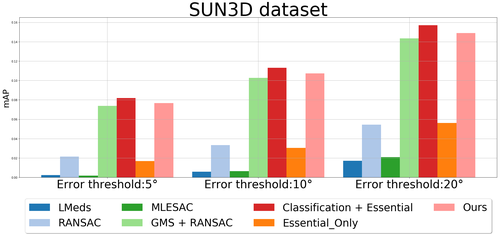}
	\label{fig:arc_sun}
\end{subfigure}
\vspace{-1em}
\caption{{\bf Keypoint matching results.} We report the accuracy of the estimated relative pose in terms of the mean Average Precision (mAP) measure of~\cite{Yi18a}. (a) Results for the SUN3D dataset. (b) Results for the dataset of \cite{Strecha08b}. Our method performs on par with the state-of-the-art method of~\cite{Yi18a}, denoted as ``Classification + Essential'', without the need of any pre-training. Note the significant performance gap between ``Essential\_Only'', which utilizes eigendecomposition directly, and our method which is eigendecomposition-free.}
\label{fig:key_quantitive}
\end{figure*}
\subsubsection{Denoising on Synthetic Data}
\label{sec:pnp_denoise}

We further evaluate the performance of our method on the denoising task for PnP.
We keep the same experimental setup as for outlier removal, except we no longer have outliers, but instead an additive Gaussian noise on the 2D coordinates. We apply a zero-mean Gaussian noise, with varying standard deviation between $5$ and $50$ pixels.
As in the ellipse-fitting case, to focus on denoising only, the network only outputs the displacements $\Delta{\bx_i}$, not the weights $\bW$, and we set $\alpha=0$ and $\beta=0$ in Eq.~\ref{eq:ell_comb} and empirically set $\gamma=10^{-2}$. As in the case of outlier rejection, we run the experiment 100 times and report the average.

We summarize the results of this experiment in Fig.~\ref{fig:pnp_denoise}. When the noise level is small, all compared methods give reasonable results. However, as the level of noise increases, our method remains accurate, whereas the others start failing.

\subsubsection{Simultaneous Outlier Removal and Denoising}

We now tackle the case where both noise and outliers are present, using synthetic data. This is a combination of the two cases in Sections~\ref{sec:pnp_outlier} and~\ref{sec:pnp_denoise}.

For the loss function in Eq.~\ref{eq:ell_comb}, we empirically set $\alpha=10$, $\beta=5\times10^{-2}$ and $\gamma=10^{-2}$.
We follow two different evaluation scenarios to demonstrate the performance in terms of outlier rejection and denoising. To this end, we vary either the number of outliers or the degree of noise, while keeping the other as a moderate value. Specifically, to demonstrate the effect of outliers at test time, we fix the image noise to be 20 pixels and vary the number of outliers from 10 to 150, out of 200 matches. To evaluate the effect of image noise, we fix the number of outliers to be 50, out of 200 matches, and vary the image noise from 5 to 50 pixels. We run the each experiment 100 times and report the average.

\Fig{pnp_combination} summarizes the results of these experiments.
In \Fig{pnp_combination} (a) and (b), we show the results for a varying number of outliers. Note that, even in the presence of noise, our method outperforms all others significantly.
This is also true when varying the noise level, as evidenced by \Fig{pnp_combination} (c) and (d).

\subsection{Wide-baseline stereo}

To evaluate our method on a real-world problem, we use the SUN3D dataset~\cite{Xiao13}. For a fair comparison, we trained our network on the same data as in~\cite{Yi18a}, that is, the ``brown-bm-3-05'' sequence, and evaluate it on the test sequences used for testing in~\cite{Ummenhofer17,Yi18a}. Additionally, to show that our method is not overfitting, we also test on a completely different dataset, the ``fountain-P11'' and ``Herz-Jesus-P8'' sequences of~\cite{Strecha08b}.

We follow the evaluation protocol of~\cite{Yi18a}, which constitutes the state-of-the-art in keypoint matching, and only change the loss function to our own loss of Eq.~\ref{eq:modelFit}. We use $\alpha=10$ and $\beta = 10^{-3}$, which we empirically found to work well for 2D-to-2D keypoint matching. We compare our method against our previous work~\cite{Yi18a}, both in its original implementation that involves minimizing a classification loss first and  then without that initial step, which we denote as ``Essential\_Only''. The latter is designed to show how critical the initial classification-based minimization of~\cite{Yi18a} is. In addition, we also compare against standard RANSAC~\cite{Cantzler}, LMeds~\cite{Simpson97}, MLESAC~\cite{Torr00}, and GMS~\cite{Bian17} to provide additional reference points. We do this in terms of the performance metric used in~\cite{Yi18a} and referred to as mean Average Precision (mAP). This metric is computed by observing the ratio of accurately recovered poses given a certain maximum threshold, and taking the area under the curve of this graph. We tuned the parameters of the baselines, so as to obtain the best results. For RANSAC, we used the threshold employed in~\cite{Yi18a} of 0.01.

We summarize the results in Figs.~\ref{fig:key_qualitative} and~\ref{fig:key_quantitive}.
Our approach performs on par with~\cite{Yi18a}, the state-of-the-art method for keypoint matching, and outperforms all the other baselines, without the need of any pre-training. Importantly, ``Essential\_Only'' severely underperforms and even often fails completely. In short, instead of having to find a workaround to the eigenvector switching problem as in~\cite{Yi18a}, we can directly optimize our objective function, which is far more generally applicable.
As discussed previously in Section~\ref{sec:motivation}, the workaround in~\cite{Yi18a} converges to a solution that depends on the pre-defined threshold for the inliers, set heuristically by the user, and is not applicable to other problems involving additive noise.
By contrast, our method can simply discover the inliers automatically while training, thanks to the second term in Eq.~\ref{eq:general}, and is also applicable to removal of additive noise.

In the top row of Fig.~\ref{fig:key_qualitative}, we compare the correspondences classified as inliers by our method to those of RANSAC on image pairs from the dataset of~\cite{Strecha08b} and SUN3D, respectively. Note that even the correspondences that are misclassified as inliers by our approach are very close to being inliers. By contrast, RANSAC yields much larger errors.
In the middle row of Fig.~\ref{fig:key_qualitative}, we show the epipolar geometry of the outliers, denoted as red lines. These lines were obtained using the estimated essential matrix in the top figure, and the ground-truth one in the bottom figure. Note that, while they may seem senseless to the human eye, the outliers lie very close to the estimated epipolar lines, and sometimes even the ground-truth ones. This is mainly because, with wide-baseline stereo, the depth of each point cannot be recovered.

\section{Conclusion}

We have introduced a novel approach to training deep networks that rely on losses computed from an eigenvector corresponding to a zero eigenvalue of a matrix defined by the network's output. Our loss does not suffer from the numerical instabilities of analytical differentiation of eigendecomposition, and converges to the correct solution much faster. 
We have demonstrated the  effectiveness of our method on two main tasks, outlier removal and denoising, with applications to keypoint matching, PnP and ellipse fitting. In all cases, our new loss has allowed us to achieve state-of-the-art results.

In the future, we will investigate extensions of our approach to handle non-zero eigenvalues, in particular the case of maximum eigenvalue. Furthermore, we hope that our work will contribute to imbuing Deep Learning techniques with traditional Computer Vision knowledge, thus avoiding discarding decades of valuable research, and favoring the development more principled frameworks.

\bibliographystyle{IEEEtran}
\bibliography{short,vision,optim,learning}

\begin{thebibliography}{10}
\providecommand{\url}[1]{#1}
\csname url@samestyle\endcsname
\providecommand{\newblock}{\relax}
\providecommand{\bibinfo}[2]{#2}
\providecommand{\BIBentrySTDinterwordspacing}{\spaceskip=0pt\relax}
\providecommand{\BIBentryALTinterwordstretchfactor}{4}
\providecommand{\BIBentryALTinterwordspacing}{\spaceskip=\fontdimen2\font plus
\BIBentryALTinterwordstretchfactor\fontdimen3\font minus
  \fontdimen4\font\relax}
\providecommand{\BIBforeignlanguage}[2]{{%
\expandafter\ifx\csname l@#1\endcsname\relax
\typeout{** WARNING: IEEEtran.bst: No hyphenation pattern has been}%
\typeout{** loaded for the language `#1'. Using the pattern for}%
\typeout{** the default language instead.}%
\else
\language=\csname l@#1\endcsname
\fi
#2}}
\providecommand{\BIBdecl}{\relax}
\BIBdecl

\bibitem{Zhang95c}
Z.~Zhang, ``{Parameter Estimation Techniques: A Tutorial with Application to
  Conic Fitting},'' \emph{IVC}, vol.~15, no.~1, pp. 59--76, 1997.

\bibitem{Kanatani16}
K.~Kanatani, Y.~Sugaya, and Y.~Kanazawa, ``{Ellipse Fitting for Computer
  Vision: Implementation and Applications},'' \emph{Synthesis Lectures on
  Computer Vision}, pp. 1--141, 2016.

\bibitem{Hartley00}
R.~Hartley and A.~Zisserman, \emph{{Multiple View Geometry in Computer
  Vision}}.\hskip 1em plus 0.5em minus 0.4em\relax Cambridge University Press,
  2000.

\bibitem{Lepetit09}
V.~Lepetit, F.~Moreno-Noguer, and P.~Fua, ``{EPnP: An Accurate O(n) Solution to
  the PnP Problem},'' \emph{IJCV}, 2009.

\bibitem{Ferraz14}
L.~Ferraz, X.~Binefa, and F.~Moreno-Noguer, ``{Very Fast Solution to the {PnP}
  Problem with Algebraic Outlier Rejection},'' in \emph{CVPR}, 2014, pp.
  501--508.

\bibitem{Zheng13}
Y.~Zheng, Y.~Kuang, S.~Sugimoto, K.~Åström, and M.~Okutomi, ``{Revisiting the
  {PnP} Problem: A Fast, General and Optimal Solution},'' in \emph{ICCV}, 2013.

\bibitem{Garro12}
V.~Garro, F.~Crosilla, and A.~Fusiello, ``{Solving the PnP Problem with
  Anisotropic Orthogonal Procrustes Analysis},'' in \emph{3DPVT}, 2012, pp.
  262--269.

\bibitem{Li12c}
S.~Li, C.~Xu, and M.~Xie, ``{A Robust {O}(n) Solution to the
  Perspective-N-Point Problem},'' \emph{PAMI}, pp. 1444--1450, 2012.

\bibitem{Yi18a}
K.~M. Yi, E.~Trulls, Y.~Ono, V.~Lepetit, M.~Salzmann, and P.~Fua, ``{Learning
  to Find Good Correspondences},'' in \emph{CVPR}, 2018.

\bibitem{Ranftl18}
R.~Ranftl and V.~Koltun, ``{Deep Fundamental Matrix Estimation},'' in
  \emph{ECCV}, 2018.

\bibitem{Papadopoulo00}
T.~Papadopoulo and M.~Lourakis, ``Estimating the jacobian of the singular value
  decomposition: Theory and applications,'' in \emph{ECCV}, 2000, pp. 554--570.

\bibitem{Giles08}
M.~Giles, ``{Collected Matrix Derivative Results for Forward and Reverse Mode
  Algorithmic Differentiation},'' in \emph{Advances in Automatic
  Differentiation}, 2008, pp. 35--44.

\bibitem{Ionescu15}
C.~Ionescu, O.~Vantzos, and C.~Sminchisescu, ``{Matrix Backpropagation for Deep
  Networks with Structured Layers},'' in \emph{CVPR}, 2015.

\bibitem{Tensorflow}
M.~Abadi, P.~Barham, J.~Chen, Z.~Chen, A.~Davis, J.~Dean, M.~Devin,
  S.~Ghemawat, G.~Irving, M.~Isard, M.~Kudlur, J.~Levenberg, R.~Monga,
  S.~Moore, D.~Murray, B.~Steiner, P.~Tucker, V.~Vasudevan, P.~Warden,
  M.~Wicke, Y.~Yu, and X.~Zheng, ``{Tensorflow: A System for Large-Scale
  Machine Learning},'' in \emph{USENIX Conference on Operating Systems Design
  and Implementation}, 2016, pp. 265--283.

\bibitem{PyTorch}
A.~Paszke, S.~Gross, S.~Chintala, G.~Chanan, E.~Yang, Z.~Devito, Z.~Lin,
  A.~Desmaison, L.~Antiga, and A.~Lerer, ``{Automatic Differentiation in
  Pytorch},'' in \emph{NIPS}, 2017.

\bibitem{Dang18a}
Z.~Dang, K.~M. Yi, Y.~Hu, F.~Wang, P.~Fua, and M.~Salzmann,
  ``{Eigendecomposition-Free Training of Deep Networks with Zero
  Eigenvalue-Based Losses},'' in \emph{ECCV}, 2018.

\bibitem{Jaderberg15}
M.~Jaderberg, K.~Simonyan, A.~Zisserman, and K.~Kavukcuoglu, ``{Spatial
  Transformer Networks},'' in \emph{NIPS}, 2015, pp. 2017--2025.

\bibitem{Handa16}
A.~Handa, M.~Bloesch, V.~Patraucean, S.~Stent, J.~McCormac, and A.~Davison,
  ``{Gvnn: Neural Network Library for Geometric Computer Vision},'' in
  \emph{ECCV}, 2016.

\bibitem{Murray16}
I.~Murray, ``{Differentiation of the Cholesky Decomposition},'' \emph{arXiv
  Preprint}, 2016.

\bibitem{Fitzgibbon99}
A.~Fitzgibbon, M.~Pilu, and R.~Fisher, ``{Direct Least-Squares Fitting of
  Ellipses},'' \emph{PAMI}, vol.~21, no.~5, pp. 476--480, May 1999.

\bibitem{Halir98}
R.~Hal{\i}r and J.~Flusser, ``{Numerically Stable Direct Least Squares Fitting
  of Ellipses},'' in \emph{International Conference in Central Europe on
  Computer Graphics and Visualization}, 1998.

\bibitem{Rousseeuw87}
P.~Rousseeuw and A.~Leroy, \emph{{Robust Regression and Outlier
  Detection}}.\hskip 1em plus 0.5em minus 0.4em\relax Wiley, 1987.

\bibitem{Kneip11}
L.~Kneip, D.~Scaramuzza, and R.~Siegwart, ``{A Novel Parametrization of the
  Perspective-Three-Point Problem for a Direct Computation of Absolute Camera
  Position and Orientation},'' in \emph{CVPR}, 2011, pp. 2969--2976.

\bibitem{Brachmann16b}
E.~Brachmann, A.~Krull, S.~Nowozin, J.~Shotton, F.~Michel, S.~Gumhold, and
  C.~Rother, ``{DSAC -- Differentiable RANSAC for Camera Localization},''
  \emph{ARXIV}, 2016.

\bibitem{Longuet-Higgins81}
H.~Longuet-Higgins, ``{A Computer Algorithm for Reconstructing a Scene from Two
  Projections},'' \emph{Nature}, vol. 293, pp. 133--135, 1981.

\bibitem{Hartley97}
R.~Hartley, ``{In Defense of the Eight-Point Algorithm},'' \emph{PAMI},
  vol.~19, no.~6, pp. 580--593, June 1997.

\bibitem{Nister03}
D.~Nister, ``{An Efficient Solution to the Five-Point Relative Pose Problem},''
  in \emph{CVPR}, June 2003.

\bibitem{Fischler81}
M.~Fischler and R.~Bolles, ``{Random Sample Consensus: A Paradigm for Model
  Fitting with Applications to Image Analysis and Automated Cartography},''
  \emph{Communications ACM}, vol.~24, no.~6, pp. 381--395, 1981.

\bibitem{Torr00}
P.~Torr and A.~Zisserman, ``{{MLESAC}: A New Robust Estimator with Application
  to Estimating Image Geometry},'' \emph{CVIU}, vol.~78, pp. 138--156, 2000.

\bibitem{Bian17}
J.~Bian, W.~Lin, Y.~Matsushita, S.~Yeung, T.~Nguyen, and M.~Cheng, ``{GMS:
  Grid-Based Motion Statistics for Fast, Ultra-Robust Feature
  Correspondence},'' in \emph{CVPR}, 2017.

\bibitem{Raguram13}
R.~Raguram, O.~Chum, M.~Pollefeys, J.~Matas, and J.-M. Frahm, ``{USAC: A
  Universal Framework for Random Sample Consensus},'' \emph{PAMI}, vol.~35,
  no.~8, pp. 2022--2038, 2013.

\bibitem{Zamir16}
A.~R. Zamir, T.~Wekel, P.~Agrawal, J.~Malik, and S.~Savarese, ``{Generic 3D
  Representation via Pose Estimation and Matching},'' in \emph{ECCV}, 2016.

\bibitem{Ummenhofer17}
B.~Ummenhofer, H.~Zhou, J.~Uhrig, N.~Mayer, E.~Ilg, A.~Dosovitskiy, and
  T.~Brox, ``{Demon: Depth and Motion Network for Learning Monocular Stereo},''
  in \emph{CVPR}, 2017.

\bibitem{Huang17a}
Z.~Huang and L.~V. Gool, ``{A Riemannian Network for SPD Matrix Learning},'' in
  \emph{AAAI}, 2017, p.~6.

\bibitem{Huang17b}
Z.~Huang, C.~Wan, T.~Probst, and L.~V. Gool, ``{Deep Learning on Lie Groups for
  Skeleton-Based Action Recognition},'' in \emph{CVPR}, 2017, pp. 6099--6108.

\bibitem{Law17}
M.~Law, R.~Urtasun, and R.~S. Zemel, ``{Deep Spectral Clustering Learning},''
  in \emph{ICML}, 2017, pp. 1985--1994.

\bibitem{Strecha08b}
C.~Strecha, W.~Hansen, L.~{Van~Gool}, P.~Fua, and U.~Thoennessen, ``{On
  Benchmarking Camera Calibration and Multi-View Stereo for High Resolution
  Imagery},'' in \emph{CVPR}, 2008.

\bibitem{Kingma15}
D.~Kingma and J.~Ba, ``{Adam: {A} Method for Stochastic Optimisation},'' in
  \emph{ICLR}, 2015.

\bibitem{Schonemann66}
P.~Sch{\"o}nemann, ``{A Generalized Solution of the Orthogonal Procrustes
  Problem},'' \emph{Psychometrika}, vol.~31, no.~1, pp. 1--10, 1966.

\bibitem{Zhang98}
Z.~Zhang, ``{Determining the Epipolar Geometry and Its Uncertainty: A
  Review},'' \emph{IJCV}, vol.~27, no.~2, pp. 161--195, 1998.

\bibitem{Gould16}
S.~Gould, B.~Fernando, A.~Cherian, P.~Anderson, R.~S. Cruz, and E.~Guo, ``{On
  differentiating parameterized argmin and argmax problems with application to
  bi-level optimization},'' \emph{arXiv preprint arXiv:1607.05447}, 2016.

\bibitem{Crivellaro18}
A.~Crivellaro, M.~Rad, Y.~Verdie, K.~M. Yi, P.~Fua, and V.~Lepetit, ``{Robust
  3D Object Tracking from Monocular Images Using Stable Parts},'' \emph{PAMI},
  2018.

\bibitem{Heinly15}
J.~Heinly, J.~Schoenberger, E.~Dunn, and J.-M. Frahm, ``{Reconstructing the
  World in Six Days},'' in \emph{CVPR}, 2015.

\bibitem{Wu13}
C.~Wu, ``{Towards Linear-Time Incremental Structure from Motion},'' in
  \emph{3DV}, 2013.

\bibitem{Xiao13}
J.~Xiao, A.~Owens, and A.~Torralba, ``{SUN3D: A Database of Big Spaces
  Reconstructed Using SfM and Object Labels},'' in \emph{ICCV}, 2013.

\bibitem{Cantzler}
H.~Cantzler, ``{Random Sample Consensus (RANSAC)},'' 2005, cVonline.

\bibitem{Simpson97}
D.~Simpson, ``{Introduction to Rousseeuw (1984) Least Median of Squares
  Regression},'' in \emph{Breakthroughs in Statistics}.\hskip 1em plus 0.5em
  minus 0.4em\relax Springer, 1997, pp. 433--461.

\end{thebibliography}
\vspace{-1.3cm}
\begin{small}
\def \bioSpacing {-1.7cm}
\begin{IEEEbiography}[{\includegraphics[width=1in,height=1.25in,clip,keepaspectratio]{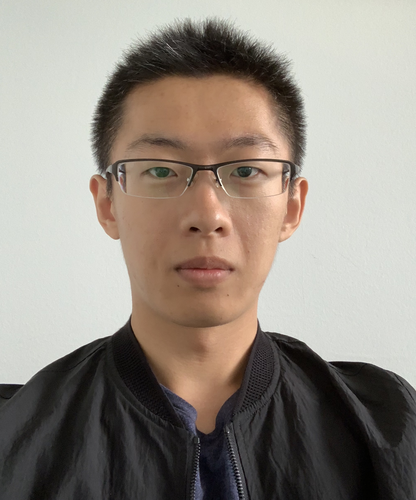}}]{Zheng Dang}
	received the B.S. degree in Automation from Northwestern Polytechnical University in 2014. Currently he is a Ph.D. candidate and pursuing his Ph.D. degree in National Engineering Laboratory for Visual Information Processing and Application, Xi’an Jiaotong University, Xi'an, China. His research interests at how to bridge the gap between geometry and deep learning.
\end{IEEEbiography}
\vspace{\bioSpacing}
\begin{IEEEbiography}[{\includegraphics[width=1in,height=1.25in,clip,keepaspectratio]{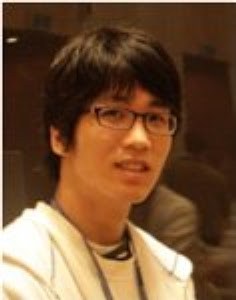}}]{Kwang Moo Yi}
	 is an assistant professor in the Department of Computer Science at the University of Victoria. Prior to UVic, he worked as a post-doctoral researcher in the Computer Vision Lab at Swiss Federal Institue of Technology at Lausanne (EPFL), working with professor Pascal Fua, and professor Vincent Lepetit. He received his Ph.D. from Seoul National University, under the supervision of professor Jin Young Choi. He also received his B.Sc. from the same institution. His research interests lie in various subjects in Computer Vision including, Learning-based methods for Visual Geometry, Learning Local Features, Augmented Reality, and Visual Surveillance.
\end{IEEEbiography}
\vspace{\bioSpacing}
\begin{IEEEbiography}[{\includegraphics[width=1in,height=1.25in,clip,keepaspectratio]{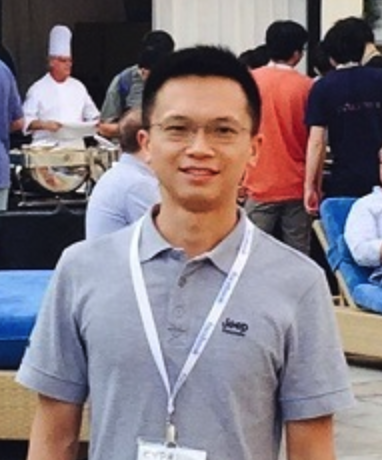}}]{Yinlin Hu}
	received his M.S. and Ph.D. degrees in Communication and Information Systems from Xidian University, China, in 2011 and 2017 respectively. Before he started to pursue the Ph.D. degree, he was a senior algorithm engineer and technical leader in Zienon, LLC. He is currently a post-doc in the CVLAB of EPFL. His research interests include optical flow, 6D pose estimation, and geometry-based learning methods.
\end{IEEEbiography}
\vspace{\bioSpacing}
\begin{IEEEbiography}[{\includegraphics[width=1in,height=1.25in,clip,keepaspectratio]{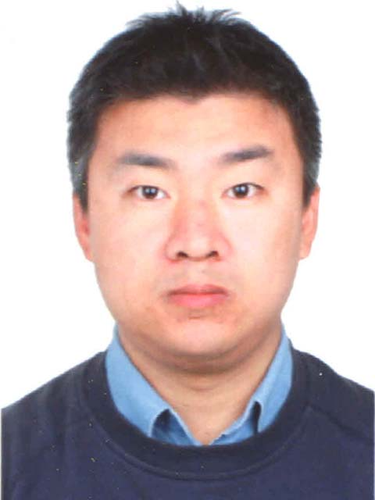}}]{Fei Wang}
	received the B.S. degree in electrical and information engineering from Northwest University, M.S. degree in communication and information system from Xi'an Institute of Optics and Precision Mechanics, Chinese Academy of Sciences, Ph.D. degree in Control science and Engineering from Xi’an Jiaotong University (XJTU), Xi’an, China, in 1998, 2002 and 2009 respectively. Currently he is a professor in Xi’an Jiaotong University. His research interests include image processing, computer vision and intelligent system.
\end{IEEEbiography}
\vspace{\bioSpacing}
\begin{IEEEbiography}[{\includegraphics[width=1in,height=1.25in,clip,keepaspectratio]{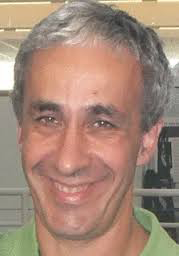}}]{Pascal Fua}
	is a Professor of Computer Science at EPFL, Switzerland. His research interests include shape and motion reconstruction from images, analysis of microscopy images, and Augmented Reality. He is an IEEE Fellow and has been an Associate Editor of the IEEE journal Transactions for Pattern Analysis and Machine Intelligence.
\end{IEEEbiography}
\vspace{\bioSpacing}
\begin{IEEEbiography}[{\includegraphics[width=1in,height=1.25in,clip,keepaspectratio]{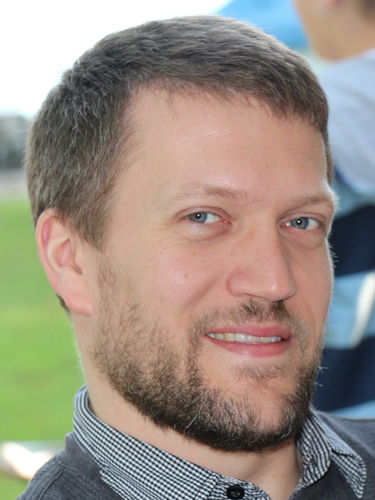}}]{Mathieu Salamann}
	is a Senior Researcher at EPFL. Previously, he was a Senior Researcher and Research Leader in NICTA’s computer vision research group, a Research Assistant Professor at TTI-Chicago, and a postdoctoral fellow at ICSI and EECS at UC Berkeley. He obtained his PhD in Jan. 2009 from EPFL. His research interests lie at the intersection of machine learning and geometry for computer vision.
\end{IEEEbiography}
\end{small}
\end{document}